\documentclass[runningheads]{llncs}

\usepackage{eccv}

\usepackage{eccvabbrv}

\usepackage{graphicx}
\usepackage{amssymb}
\usepackage{amssymb}
\usepackage{dsfont}
\usepackage{epstopdf}
\usepackage{listings}
\usepackage{amsmath}
\usepackage{pifont}
\usepackage{enumitem}
\usepackage{multirow}
\usepackage{soul}
\usepackage{booktabs}
\usepackage{marvosym}
\usepackage{makecell}
\usepackage{stfloats}
\usepackage{makecell}
\usepackage{xspace}
\usepackage{url}
\usepackage{wrapfig}
\usepackage{placeins}
\usepackage{float}
\usepackage{multirow}
\usepackage{makecell}
\usepackage{color}
\usepackage[table, dvipsnames]{xcolor}
\definecolor{blue}{HTML}{5a8ad2}
\definecolor{red}{HTML}{C55A11}
\definecolor{purple}{HTML}{C686EE}
\definecolor{gray}{HTML}{E8E8E8}
\definecolor{teaserred}{HTML}{C14D4D}
\definecolor{hr}{HTML}{FFC28F}
\definecolor{lr}{HTML}{ABCBE3}
\usepackage[accsupp]{axessibility}  

\usepackage{hyperref}

\usepackage{orcidlink}

\begin{document}

\title{LinCa: Accelerating Diffusion Models via Learnable Decomposed Feature Caching}
\titlerunning{LinCa: Learnable Decomposed Feature Caching}


\author{
  \textbf{Jinshan Liu}\inst{1,3}\textsuperscript{$*$} \and
  \textbf{Haoran Qin}\inst{1,4}\textsuperscript{$*$} \and
  \textbf{Xiaobing Tu}\inst{2} \and
  \textbf{Jiacheng Liu}\inst{1} \and
  \textbf{Jiahui Hu}\inst{1,5} \and
  \textbf{Zhengan Yan}\inst{1} \and
  \textbf{Yukun Xie}\inst{1,3} \and
  \textbf{Kerui Shen}\inst{1,3} \and
  \textbf{Jinkui Ren}\inst{2} \and
  \textbf{Yuqi Lin}\inst{1,6} \and
  \textbf{Xiantao Zhang}\inst{2} \and
  \textbf{Linfeng Zhang}\inst{1}\textsuperscript{$\dagger$}
}
\authorrunning{J. Liu et al.}

\institute{$^{1}$Shanghai Jiao Tong University, China \quad $^{2}$Terminal Intelligent Computing Division, Alibaba Cloud, China \quad $^{3}$Xi'an Jiaotong University, China \quad $^{4}$Shandong University, China \quad $^{5}$South China University of Technology, China \quad $^{6}$Jilin University, China}

\maketitle

\begingroup
\renewcommand{\thefootnote}{$*$}
\footnotetext{Equal contribution.}
\renewcommand{\thefootnote}{$\dagger$}
\footnotetext{Corresponding author. Email: \texttt{zhanglinfeng@sjtu.edu.cn}}
\endgroup

\begin{figure}
    \centering
    \vspace{-8mm}
    \includegraphics[width=0.9\linewidth]
    {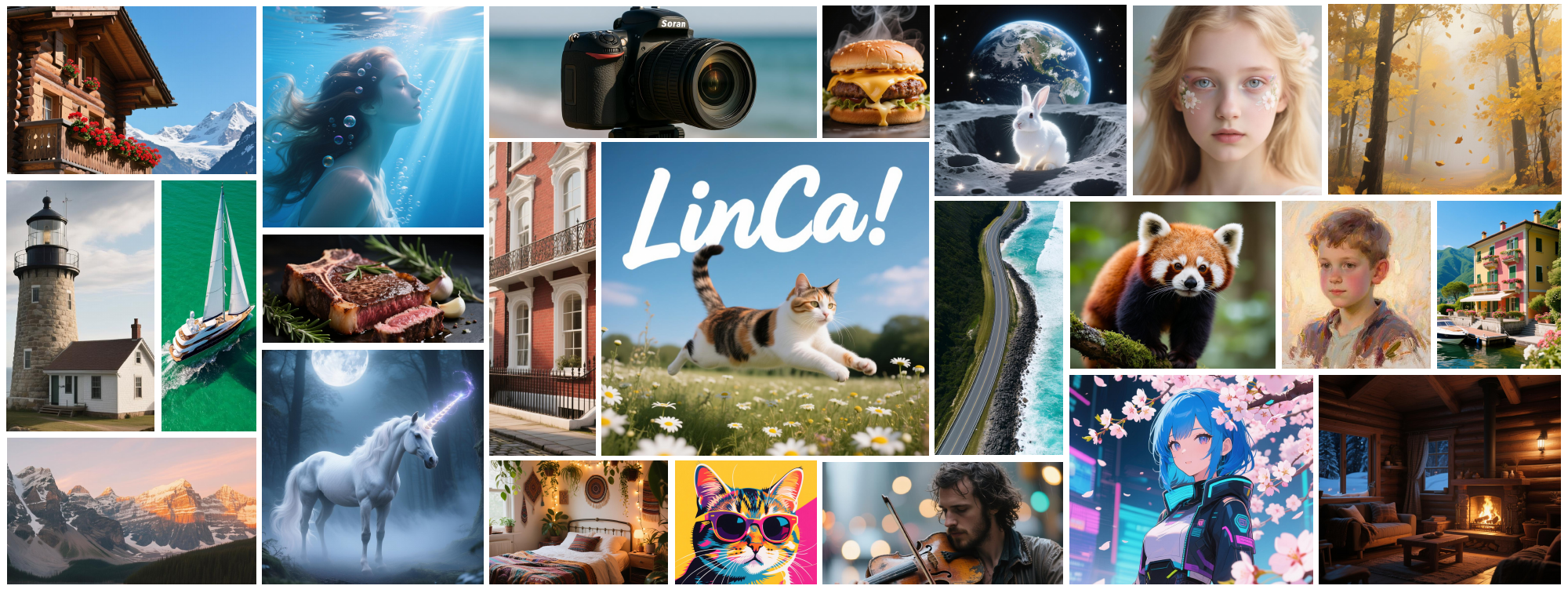}
    \captionof{figure}{Images sampled by Qwen-Image with \textit{LinCa} at \textbf{6.95}$\times$ acceleration.}
    \label{fig:head}
    \vspace{-11mm}
\end{figure}

\begin{abstract}

  Diffusion models have achieved remarkable success in image and video generation, yet the high computational cost of iterative sampling remains a critical bottleneck for practical deployment. Feature caching has emerged as a promising acceleration paradigm by reusing or predicting intermediate features across timesteps. However, existing training-free methods apply uniform prediction strategies that cannot adapt to the heterogeneous feature dynamics, causing significant quality degradation under high acceleration ratios. We propose \textbf{LinCa}, a feature caching framework based on learnable invertible networks. LinCa decomposes cached features into sub-components with distinct continuity properties via a lightweight invertible network and applies differentiated prediction orders matched to each component. The strict invertibility guarantees lossless reconstruction back to the original feature space, forming a unified Decompose-Predict-Reconstruct pipeline. By training separate predictors for different models and timestep segments, LinCa adapts to heterogeneous feature dynamics. Experiments on FLUX, Qwen-Image, and HunyuanVideo demonstrate that LinCa, with less than $0.2\%$ additional parameters, significantly outperforms existing methods and maintains near-lossless quality at $5$-$7\times$ speedup. Code: \href{https://github.com/QHR69/LinCa}{https://github.com/QHR69/LinCa}.
  
  \keywords{Diffusion Acceleration \and Learnable Invertible Network}
\end{abstract}

\section{Introduction}
\label{sec:intro}


\begin{figure}[h]
    \vspace{-5mm}
    \centering
    \includegraphics[width=0.79\linewidth]
    {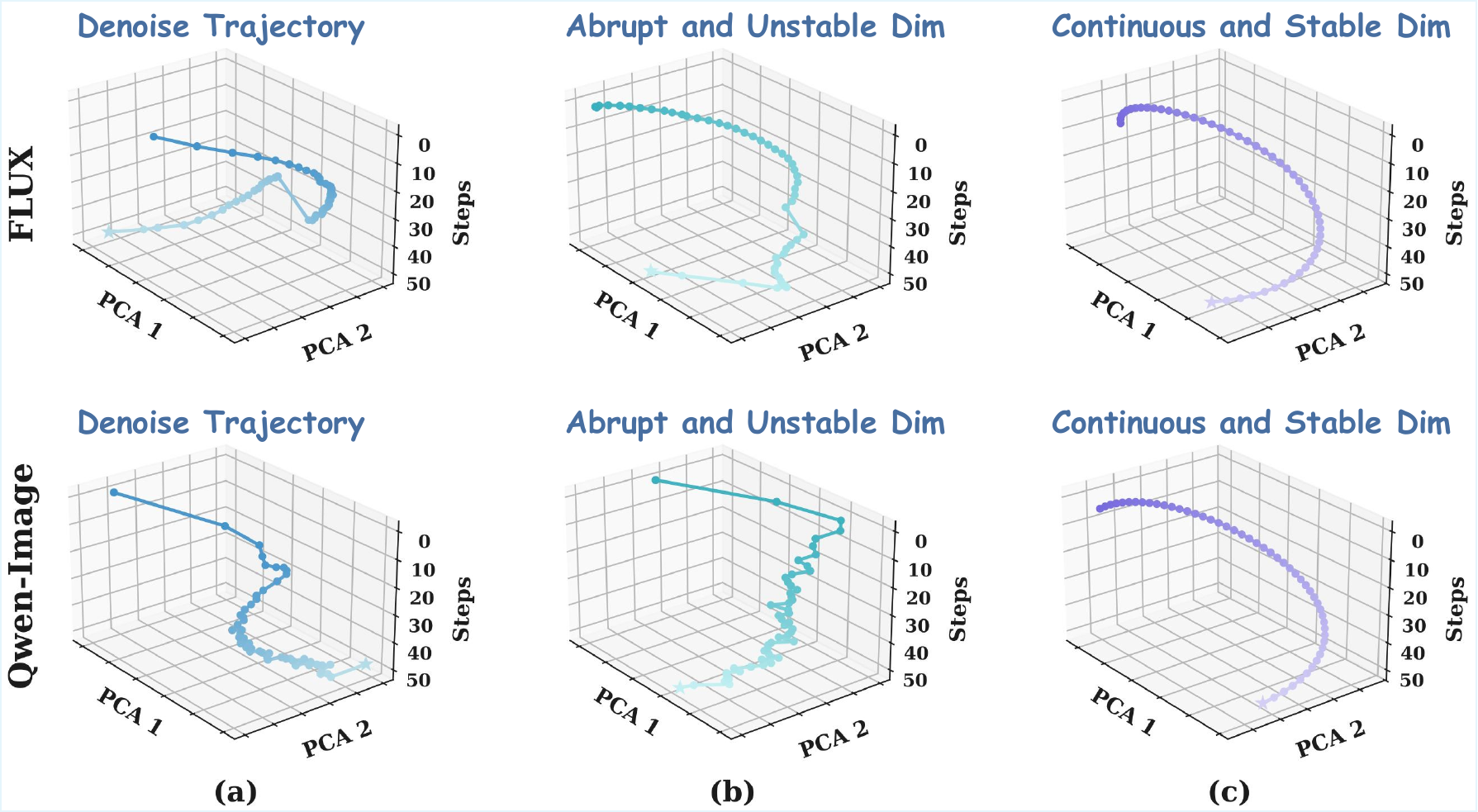}
    \captionof{figure}{\textbf{Analysis of dynamics mismatch. (a):} Denoise trajectories visualized via Principal Component Analysis (PCA) within FLUX.1-dev and Qwen-Image. Significant continuity differences exist across different models and denoising stages. \textbf{(b)-(c):} Denoise trajectories of specific dimensions visualized via PCA within FLUX.1-dev and Qwen-Image. Even within the same model and stage, significant continuity differences still exist across different feature dimensions. (b) shows dimensions with low continuity and abrupt mutations, while (c) shows dimensions with high continuity and stability.}
    \label{fig:intro}
    \vspace{-5mm}
\end{figure}

Diffusion models have achieved remarkable progress in image and video generation and editing~\cite{DM,StableDiffusion,blattmann2023SVD}. Recently, Diffusion Transformers~\cite{peebles2023dit} have become the predominant architecture for high-quality conditional generation and editing, owing to their superior scalability and modeling capacity. However, the iterative denoising sampling mechanism of Diffusion Transformers requires multi-step forward passes, and the resulting high computational cost makes efficient inference a critical bottleneck for practical deployment.

To address this challenge, two main acceleration directions have emerged: reducing the total number of sampling steps through algorithmic advances~\cite{lu2022dpm}, and lowering the per-step cost through architectural optimization~\cite{yuan2024ditfastattnattentioncompressiondiffusion, zhao2025realtimevideogenerationpyramid}. Among these, feature caching exploits the temporal consistency of hidden representations across adjacent timesteps and has become a particularly promising solution. Specifically, it performs a full model forward pass only once every $N$ timesteps and caches the intermediate features. The remaining $N{-}1$ steps directly reuse or predict features based on previous caches, thereby skipping expensive forward computation to achieve acceleration, as exemplified by recent works such as FORA, ToCa, and TaylorSeer~\cite{selvaraju2024fora,zou2024accelerating,liu2025reusingforecastingacceleratingdiffusion}. Despite this progress, we identify the following limitations in current feature caching methods.

\textbf{Cross-timestep and cross-model dynamics mismatch.} Existing feature caching methods implicitly assume that the hidden features of diffusion models follow a uniform evolution pattern throughout the entire denoising process, and thus apply a single preset caching strategy across all timesteps and all models. However, as shown in Figure~\ref{fig:intro}(a), we find that even within the same model, the feature evolution patterns differ significantly across timesteps. Some timestep segments exhibit stable and continuous trajectories, while others display poor continuity accompanied by abrupt mutations. Across different models, the discrepancy in feature trajectory shapes and variation patterns is more pronounced. This indicates that the denoising process is not governed by a single dynamics mechanism, and applying a uniform caching rule across all timesteps and models inevitably introduces cumulative errors in certain scenarios.

\textbf{Cross-dimension dynamics mismatch.} From a finer-grained perspective, existing methods apply a uniform prediction strategy to all feature dimensions, implicitly assuming that all dimensions share the same continuity. However, as shown in Figure~\ref{fig:intro}(b)-(c), even within the same model at the same timestep segment, different feature dimensions exhibit markedly different dynamics. Some dimensions display good stability and continuity, making them suitable for high-order polynomial prediction. Other dimensions are unstable and accompanied by abrupt mutations, making them difficult to predict directly. More critically, dimensions with different continuity properties are interleaved in the original feature space and cannot be separated by simple dimensional partitioning. This suggests that a uniform prediction strategy cannot accommodate the vastly different dynamics across feature dimensions.

To address the dynamics mismatch at both levels, this paper introduces \textbf{LinCa}, a \textbf{L}earnable \textbf{I}nvertible \textbf{N}etworks based Feature \textbf{Ca}ching acceleration framework. Through minimal parameter overhead and training cost, LinCa employs a learnable mapping to decompose cached features into sub-components with different dynamics and applying the corresponding differentiated prediction orders to each component. Unstable components are directly reused from the nearest cache, while components with good continuity are predicted via polynomial extrapolation matched to their continuity from historical caches. This learnable mapping is realized by a lightweight fully invertible network whose strict invertibility guarantees that the decomposed features can be losslessly reconstructed back to the original feature space, forming an end-to-end learnable Decompose-Predict-Reconstruct pipeline. Furthermore, we partition the denoising timesteps into multiple segments and train isomorphic but independently parameterized predictors, adapting to the distinct feature dynamics of different diffusion models and timestep segments. With a parameter count less than $0.2\%$ of the original diffusion model, LinCa requires only 100-200 pre-generated features as training data, and completes training within $1$ hour on a single GPU (12GB VRAM) without loading the diffusion model weights. During inference, LinCa achieves nearly the same speed as training-free methods, further enhancing its practical value.

LinCa delivers robust and efficient feature prediction across diverse tasks and architectures. It achieves near-lossless acceleration of $\textbf{5.51}\times$ on FLUX.1-dev, $\textbf{6.95}\times$ on Qwen-Image, $\textbf{7.08}\times$ on Qwen-Image-Edit, and $\textbf{5.50}\times$ on HunyuanVideo (Figure~\ref{fig:head}). Moreover, it is also fully compatible with distillation or quantization with higher acceleration ratio and strong image quality. In summary, our main contributions are as follows:

\begin{itemize}[leftmargin=10pt,topsep=-2pt,label=\textbf{\textbullet}]

    \item \textbf{Heterogeneous Feature Dynamics:} We reveal that the hidden features of diffusion models exhibit significant dynamics mismatch across timesteps and models, and that different feature dimensions display markedly different evolution patterns, highlighting the importance of adaptively matching feature dynamics through learnable mappings.
    \item \textbf{LinCa Framework:} We propose \textbf{LinCa}, a feature caching framework that decomposes cached features via a lightweight invertible network and applies differentiated prediction orders. By training separately for different models across different timestep segments, LinCa adapts to heterogeneous feature dynamics with minimal training overhead.
    \item \textbf{Superior Performance:} We evaluate LinCa on diverse architectures and tasks including FLUX.1-dev, Qwen-Image, Qwen-Image-Edit, HunyuanVideo and even distilled or quantized models. Across all settings, LinCa significantly outperforms existing training-free methods at the same acceleration ratio and maintains near-lossless generation quality under high speedup.
\end{itemize}

\section{Related Works}

Diffusion models~\cite{sohldickstein2015deepunsupervisedlearningusing,ho2020DDPM} have become the dominant framework for high-fidelity image and video generation. The Diffusion Transformer (DiT)~\cite{peebles2023dit} further advanced this field through superior scalability and expressive capacity, becoming foundational for large-scale visual generation systems~\cite{chen2023pixartalpha,chen2024pixartsigma,opensora,yang2025cogvideox,ma2026group,ma2024followpose,ma2025followcreation,ma2026fastvmt,ma2025followyourmotion}. Nevertheless, the iterative nature of the sampling process remains the primary inference bottleneck, and current acceleration efforts focus on two complementary directions: reducing sampling steps and accelerating the denoising network.

\subsection{Sampling Timestep Reduction}

DDIM~\cite{songDDIM} introduced deterministic few-step sampling, further refined by the DPM-Solver series~\cite{lu2022dpm,lu2022dpm++} through high-order ODE solvers. Rectified Flow~\cite{refitiedflow} shortens the transport path via optimal transport, while knowledge distillation~\cite{salimans2022progressive,meng2022on} compresses long sampling trajectories into compact student generators. Consistency Models~\cite{song2023consistency} further enable single-step synthesis by learning a direct noise-to-data mapping. Although effective, these methods typically require redesigning sampling algorithms or retraining the diffusion model, limiting their applicability to pre-trained diffusion models.

\subsection{Denoising Network Acceleration}

Another acceleration direction focuses on reducing the computational cost of each forward pass, primarily through model compression and feature caching.

\textbf{Model Compression-based Acceleration.} Model compression reduces inference overhead through structured pruning~\cite{structural_pruning_diffusion,zhu2024dipgo}, quantization~\cite{shang2023post,kim2025ditto}, distillation~\cite{li2024snapfusion}, and token merging or pruning~\cite{bolya2023token,kim2024tofu,zhang2024tokenpruningcachingbetter,zhang2025sito,cheng2025catpruningclusterawaretoken}. However, these methods often necessitate additional training or fine-tuning to maintain generation quality, and aggressive compression may compromise robustness~\cite{li2024snapfusion}.

\textbf{Feature Caching-based Acceleration.} Feature caching avoids redundant computation by reusing activations across timesteps, attracting wide attention for its ability to work without modifying model structure and typically in a training-free manner. Current works~\cite{cai2026lesa, liu2025survey} extended caching to DiT architectures: FORA~\cite{selvaraju2024fora} and $\Delta$-DiT~\cite{chen2024delta-dit} cache block outputs, while ToCa and DuCa~\cite{zou2024accelerating,zou2024DuCa} adaptively reuse tokens or regions. TaylorSeer~\cite{liu2025reusingforecastingacceleratingdiffusion} introduced polynomial extrapolation from cached features, further advanced by FoCa~\cite{zhengFoCa2025} and SpeCa~\cite{Liu2025SpeCa}. More recently, Clusca~\cite{Zheng2025Compute} reduces token redundancy via spatial clustering, while HyCa~\cite{zheng2025let} applies dimension-wise caching based on mixed ODE modeling.

However, all the above methods treat cached features as a whole, applying a uniform prediction strategy across all feature dimensions and overlooking their markedly different continuity. Meanwhile, feature dynamics differ significantly across model architectures and denoising stages, yet existing training-free methods apply the same strategy across all scenarios, leading to cumulative prediction errors and notable quality degradation under high acceleration ratios.

In contrast, \textbf{LinCa} employs invertible networks capable of lossless reconstruction to map cached features into sub-components with distinct continuity properties for differentiated order prediction, addressing the quality degradation inherent in existing caching methods. Meanwhile, data-driven per-segment training enables LinCa to adaptively accommodate the feature dynamics differences across model architectures and denoising stages with minimal training cost, significantly improving prediction performance under high acceleration ratios.

\section{Methodology}

\subsection{Preliminary}

Feature caching accelerates diffusion model inference by avoiding redundant computation across timesteps. Let $\mathbf{x}_t$ denote the hidden feature produced by the diffusion model at timestep $t$, feature caching perform a full forward pass every $N$ steps and cache the intermediate features, while estimating the remaining $N{-}1$ steps through a prediction function $f$ based on historical caches:
\begin{equation}
    \hat{\mathbf{x}}_{t-k} = f(\mathbf{x}_{t},\, \mathbf{x}_{t+N},\, \ldots), \quad k \in \{1, \ldots, N{-}1\}
\end{equation}

\begin{figure}[tbp]
    \centering
    \includegraphics[width=\linewidth]
    {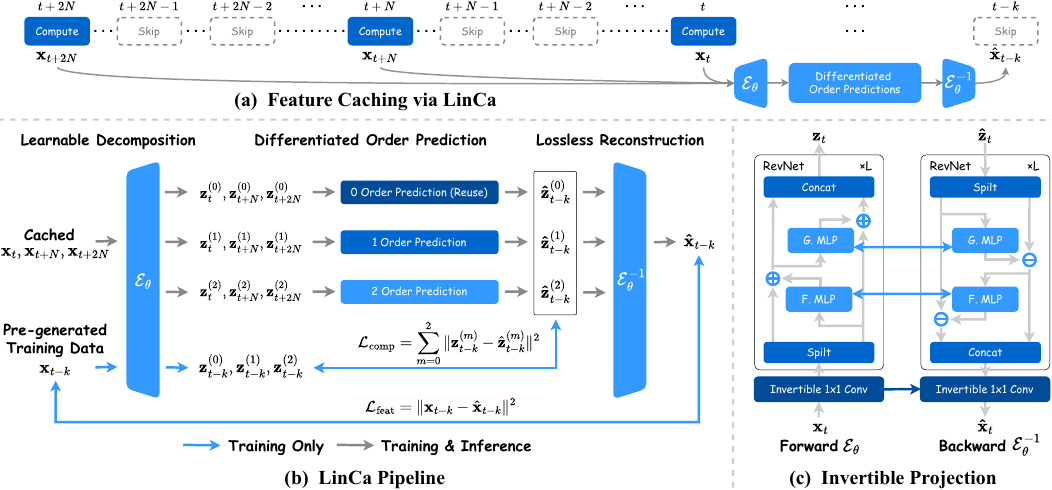}
    \captionof{figure}{\textbf{Overview of the LinCa framework.} \textbf{(a) Feature Caching via LinCa:} Feature caching caches features during computation steps and skips computation by utilizing historical cached features during prediction steps, where LinCa provides an end-to-end ``Decompose-Predict-Reconstruct'' pipeline. \textbf{(b) LinCa Pipeline:} Cached features are decomposed into sub-components with distinct continuity, predicted by differentiated order polynomial extrapolation, and losslessly reconstructed back to the original feature space. \textbf{(c) Invertible Projection:} Each invertible block consists of an invertible $1\times1$ convolution and an additive coupling layer with lightweight MLPs $F$ and $G$. The strict invertibility of each sub-layer guarantees lossless reconstruction.}

    \label{fig:overview}
    \vspace{-5mm}
\end{figure}

where $\mathbf{x}_{t}, \mathbf{x}_{t+N}, \ldots$ are previously cached features. Existing feature caching methods aim to construct an effective $f$ for accurate prediction. For instance, when $f$ is modeled as the identity mapping, it corresponds to direct reuse: $\hat{\mathbf{x}}_{t-k}=\mathbf{x}_{t}$. Alternatively, $f$ can be modeled as a Taylor expansion: $\hat{\mathbf{x}}_{t-k} = \mathbf{x}_t + \sum_{i=1}^{m} \frac{\Delta^i \mathbf{x}_t}{i! \cdot N^i}\,(-k)^i$, where $m$ is the prediction order and $\Delta^i \mathbf{x}_t$ denotes the $i^{th}$ order discrete difference. $f$ can also be modeled as Hermite interpolation based on discrete differences: $\hat{\mathbf{x}}_{t-k} = \mathbf{x}_{t} + \sum_{i=1}^{m} \alpha_i(k)\, \Delta^i \mathbf{x}_{t}$, where $\alpha_i(k)$ are interpolation coefficients derived from Hermite polynomials.

However, regardless of the form of $f$, existing methods apply a uniform prediction strategy and prediction order across all models, all timesteps, and all feature dimensions, failing to accommodate the significant dynamics differences at each of these levels and thus compromising generation quality.

\subsection{Learnable Invertible Network based Feature Caching}

In this section, we introduce the \textbf{LinCa} framework (Figure~\ref{fig:overview}). To address the dynamics mismatch at multiple levels in existing methods, LinCa is built upon two key components: (i) decomposing cached features into sub-components with distinct continuity properties through a learnable mapping and applying order-matched polynomial prediction to each sub-component, and (ii) realizing this mapping with a lightweight invertible network whose strict invertibility guarantees lossless reconstruction of the input features, trained separately for different models and timestep segments to adapt to their distinct feature dynamics.

\subsubsection{Learnable Decomposition and Differentiated Order Prediction. }

Our differentiated prediction strategy is motivated by the observation that different feature dimensions exhibit distinct dynamics. As shown in Figure~\ref{fig:intro}(b)-(c), some feature dimensions are unstable with abrupt mutations and low temporal continuity, making them difficult to predict. In contrast, other dimensions exhibit varying degrees of temporal continuity, with diffusion trajectories suitable for polynomial prediction at different orders. However, dimensions with different continuity properties are interleaved in the original feature space and cannot be separated by simple partitioning. We therefore propose to learn a differentiated decomposition strategy that regroups and separates feature dimensions with different evolution characteristics, applying prediction functions $f$ of different orders to each group.

To this end, following prior work~\cite{liu2025freqca}, we cache the cumulative residual feature at the final layer of the diffusion model. For the cached feature $\mathbf{x}_t \in \mathbb{R}^{N \times D}$ at timestep $t$, LinCa projects it through a learnable mapping $\mathcal{E}_\theta$ and decomposes it along the feature dimension into $M$ sub-components ($M=3$ in practice):
\begin{equation}
\mathbf{z}_t = \mathcal{E}_\theta(\mathbf{x}_t) = [\mathbf{z}_t^{(0)} \mid \mathbf{z}_t^{(1)} \mid \cdots \mid \mathbf{z}_t^{(M-1)}], \quad \mathbf{z}_t^{(m)} \in \mathbb{R}^{N \times d_m}
\end{equation}

where $\mathcal{E}_\theta$ is continuously optimized during training to cluster dimensions with similar continuity into corresponding sub-spaces, thereby regrouping and separating feature dimensions with different evolution characteristics. Based on the distinct dynamics of each sub-component, we apply differentiated order prediction strategies. For the $0^{th}$ order sub-component, which is unstable with weak continuity, we directly reuse the nearest cached value: $\hat{\mathbf{z}}_t^{(0)} = \mathbf{z}_{t_{\text{prev}}}^{(0)}$. For higher-order sub-components ($m \geq 1$), we predict each using $m^{th}$ order Hermite interpolation. These closed-form predictors accurately reconstruct higher-order details at negligible computational cost. Finally, all sub-components are concatenated and reconstructed back to the original feature space through the inverse mapping $\mathcal{E}_\theta^{-1}$, yielding the final predicted feature:
\begin{equation}
\hat{\mathbf{x}}_t = \mathcal{E}_\theta^{-1}([\hat{\mathbf{z}}_t^{(0)} \mid \hat{\mathbf{z}}_t^{(1)} \mid \cdots \mid \hat{\mathbf{z}}_t^{(M-1)}])
\end{equation}

\subsubsection{Lossless Mapping via Invertible Networks. }

The above Decompose-Predict-Reconstruct pipeline requires $\mathcal{E}_\theta$ to be both learnable and capable of losslessly reconstructing the predicted sub-components back to the original feature space through its inverse $\mathcal{E}_\theta^{-1}$. To achieve this, following~\cite{gomez2017reversible}, we design a lightweight invertible network whose architecture guarantees strict mathematical invertibility, ensuring that the reconstruction process introduces no additional loss.

Concretely, $\mathcal{E}_\theta$ consists of $L$ stacked invertible blocks. Each block contains two invertible sub-layers: an invertible $1\times1$ convolution (parameterized by an orthogonal matrix $\mathbf{W}$) for channel mixing, followed by an additive coupling layer that evenly splits the features along the feature dimension into $\mathbf{u}_1, \mathbf{u}_2$ and applies lightweight MLPs $F, G$:
\begin{equation}
\mathbf{v}_1 = \mathbf{u}_1 + F(\mathbf{u}_2), \quad \mathbf{v}_2 = \mathbf{u}_2 + G(\mathbf{v}_1)
\end{equation}
The outputs $\mathbf{v}_1, \mathbf{v}_2$ are concatenated along the feature dimension. During reconstruction, the corresponding inverse mapping is executed: first recovering the pre-coupling components via subtraction $\mathbf{u}_2 = \mathbf{v}_2 - G(\mathbf{v}_1)$, $\mathbf{u}_1 = \mathbf{v}_1 - F(\mathbf{u}_2)$, then restoring the channel mixing through $\mathbf{W}^{-1}$. Since each sub-layer is strictly invertible, the entire network satisfies $\mathcal{E}_\theta^{-1} \circ \mathcal{E}_\theta = \mathbf{I}$, introducing no additional error and thus guaranteeing the highest reconstruction quality.

To adapt to the feature dynamics of different timestep segments, we partition the denoising process into $S$ segments and independently train an isomorphic but separately parameterized predictor $\mathcal{E}_\theta^{(s)}$ for each segment. LinCa requires only 100-200 pre-generated image features and completes training within one hour on a single GPU (12GB VRAM) without loading the diffusion model weights, adapting to the feature dynamics of different diffusion models at minimal cost.

In detail, during training, we load the target feature $\mathbf{x}_{t-k}$ and historical cached features $\mathbf{x}_{t}, \mathbf{x}_{t+N}, \ldots$ from pre-generated data, and decompose them through $\mathcal{E}_\theta^{(s)}$ into sub-components $\mathbf{z}_{t-k}^{(m)}$ and $\mathbf{z}_{t}^{(m)}, \mathbf{z}_{t+N}^{(m)}, \ldots$ respectively. The historical sub-components are then used to predict the target sub-components $\hat{\mathbf{z}}_{t-k}^{(m)}$ via differentiated order extrapolation, and the predictions are reconstructed through the inverse mapping $(\mathcal{E}_\theta^{(s)})^{-1}$ to yield $\hat{\mathbf{x}}_{t-k}$. The optimization objective for each segment is:
\begin{equation}
\mathcal{L}^{(s)} = \mathcal{L}_{\text{feat}}^{(s)} + \lambda\,\mathcal{L}_{\text{comp}}^{(s)}
\end{equation}
where the end-to-end prediction loss $\mathcal{L}_{\text{feat}}^{(s)} = \|\hat{\mathbf{x}}_{t-k} - \mathbf{x}_{t-k}\|^2$ ensures the overall prediction quality in the original feature space after inverse reconstruction. The sub-component prediction loss $\mathcal{L}_{\text{comp}}^{(s)} = \sum_{m=0}^{M-1} \|\hat{\mathbf{z}}_{t-k}^{(m)} - \mathbf{z}_{t-k}^{(m)}\|^2$ directly constrains the prediction accuracy of each sub-component, driving $\mathcal{E}_\theta^{(s)}$ to cluster dimensions suitable for $m^{th}$ order prediction into the corresponding sub-space.

\section{Experiments}

\subsection{Experiment Settings}

\textbf{Model Configurations.} The experiments are conducted on four state-of-the-art diffusion-based models: the text-to-image models FLUX.1-dev~\cite{flux2024} and Qwen-Image~\cite{wu2025qwenimagetechnicalreport}, the text-to-video model HunyuanVideo~\cite{sun_hunyuan-large_2024}, and the image editing model Qwen-Image-Edit~\cite{wu2025qwenimagetechnicalreport}. To further assess compatibility with model compression techniques, we also evaluate our method on distilled or quantized models: FLUX.1-lite-8B~\cite{flux1lite_8B}, FLUX.1-schnell~\cite{flux1_schnell} and FLUX.1-dev-int8~\cite{flux1dev_torchao_int8}. All experiments are conducted on NVIDIA A100 GPUs for the FLUX series, H100 GPUs for HunyuanVideo, and H20 GPUs for Qwen-Image and Qwen-Image-Edit. More details are provided in the supplementary materials.

\textbf{Evaluation and Metrics.} For text-to-image generation, we follow the DrawBench~\cite{sahariaPhotorealisticTexttoImageDiffusion2022} protocol and evaluate all models on a fixed set of 200 prompts, assessing images using ImageReward~\cite{xuImageRewardLearningEvaluating2023} for photorealism, CLIP Score~\cite{hessel2021clipscore} for text-image alignment, and PSNR, SSIM, and LPIPS for fidelity. For text-to-video generation, we evaluate our model on VBench~\cite{VBench}, which provides multi-dimensional assessments covering motion quality, visual appearance, and semantic consistency. Regarding image editing tasks, we utilize GEdit-Bench~\cite{liu2025step1xeditpracticalframeworkgeneral} to evaluate model performance across a diverse set of edit types and prompts.

\subsection{Results on Text-to-Image Generation}


\begin{table*}[htbp]
\centering
\vspace{-5mm}
\caption{\textbf{Quantitative comparison of text-to-image generation} for FLUX.1-dev. Best results are highlighted in \textbf{bold}, and second-best are \underline{underlined}.}
\vspace{-3mm}
\resizebox{\textwidth}{!}{
\begin{tabular}{l | c  c | c  c | c | c }
    \toprule
    \rowcolor{white}
  \multirow[c]{2}{*}{\bf Method} & \multicolumn{4}{c|}{\bf Acceleration} & \multirow{2}{*} {\bf ImageReward $\uparrow$} & \multirow{2}{*}{\bf CLIP Score $\uparrow$} \\
    \cline{2-5}
      & {\bf Latency(s) $\downarrow$} & {\bf Speed $\uparrow$} & {\bf FLOPs(T) $\downarrow$}  & {\bf Speed $\uparrow$}\rule{0pt}{2ex} &  & \\
    \midrule
        $\textbf{Original: 50 steps}$  & 23.10 & 1.00$\times$ & 3719.50 & 1.00$\times$ & 0.9930 \textcolor{black!40}{\scriptsize ($+$0.0\%)} & 32.61 \textcolor{black!40}{\scriptsize ($+$0.0\%)} \\ 
        \midrule

        {$60\%$\textbf{ steps}} & 14.87 & 1.55$\times$ & 2231.70 & 1.67$\times$ & 0.9693 \textcolor{black!40}{\scriptsize ($-$2.4\%)} & 32.50 \textcolor{black!40}{\scriptsize ($-$0.3\%)} \\
        {$\Delta$$-$DiT} ($\mathcal{N}=2$)~\cite{chen2024delta-dit} & 15.96 & 1.45$\times$ & 2480.01 & 1.50$\times$ & 0.9471 \textcolor{black!40}{\scriptsize ($-$4.6\%)} & 32.46 \textcolor{black!40}{\scriptsize ($-$0.5\%)} \\
        {$\Delta$$-$DiT} ($\mathcal{N}=3$)~\cite{chen2024delta-dit} & 11.63 & 1.98$\times$ & 1686.76 & 2.21$\times$ & 0.8750 \textcolor{black!40}{\scriptsize ($-$11.9\%)} & 32.29 \textcolor{black!40}{\scriptsize ($-$1.0\%)} \\ 
        $\textbf{FORA}$ $(\mathcal{N}=3)$~\cite{selvaraju2024fora} & 9.08 & 2.54$\times$ & \underline{1320.07} & \underline{2.82}$\times$ & 0.9802 \textcolor{black!40}{\scriptsize ($-$1.3\%)} & 32.45 \textcolor{black!40}{\scriptsize ($-$0.5\%)} \\
        $\textbf{DBCache}$ $(\mathcal{F}=8,B=8)$~\cite{cache-dit@2025} & 15.05 & 1.53$\times$ & 2384.29 & 1.56$\times$ & \underline{1.0097} \textcolor{black!40}{\scriptsize ($+$1.7\%)} & 32.72 \textcolor{black!40}{\scriptsize ($+$0.3\%)} \\
        $\textbf{TaylorSeer}$ $(\mathcal{N}=3, O=2)$~\cite{liu2025reusingforecastingacceleratingdiffusion} & 8.83 & 2.61$\times$ & \underline{1320.07} & \underline{2.82}$\times$ & 1.0018 \textcolor{black!40}{\scriptsize ($+$0.9\%)} & 32.58 \textcolor{black!40}{\scriptsize ($-$0.1\%)} \\
        $\textbf{FoCa}$ $(\mathcal{N}=3)$~\cite{zhengFoCa2025} & \underline{8.35} & \underline{2.78}$\times$ & 1327.21 & 2.80$\times$ & 0.9917 \textcolor{black!40}{\scriptsize ($-$0.1\%)} & \underline{32.75} \textcolor{black!40}{\scriptsize ($+$0.4\%)} \\
        \rowcolor{gray!100}
        $\textbf{LinCa}$ $(\mathcal{N}=4)$ & \bf 7.51 & \textbf{3.08$\times$} & \bf 1120.68 & \textbf{3.32$\times$} &\bf 1.0175 \textcolor[HTML]{0f98b0}{\scriptsize ($+$2.5\%)} & \bf 32.88 \textcolor[HTML]{0f98b0}{\scriptsize ($+$0.8\%)} \\
        \midrule

        {$34\%$\textbf{ steps}} & 8.10 & 2.85$\times$ & 1264.63 & 3.13$\times$ & 0.9482 \textcolor{black!40}{\scriptsize ($-$4.5\%)} & 32.28 \textcolor{black!40}{\scriptsize ($-$1.0\%)} \\
        $\textbf{Chipmunk}$~\cite{silveria2025chipmunk} & 11.39 & 2.02$\times$ & 1505.87 & 2.47$\times$ & 0.9965 \textcolor{black!40}{\scriptsize ($+$0.4\%)} & 32.71 \textcolor{black!40}{\scriptsize ($+$0.3\%)} \\
        $\textbf{FORA}$ $(\mathcal{N}=4)$~\cite{selvaraju2024fora} & 7.28 & 3.14$\times$ & 967.91 & 3.84$\times$ & 0.9757 \textcolor{black!40}{\scriptsize ($-$1.7\%)} & 32.31 \textcolor{black!40}{\scriptsize ($-$0.9\%)} \\
        $\textbf{\texttt{ToCa}}$ $(\mathcal{N}=6)$~\cite{zou2024accelerating} & 11.76 & 1.96$\times$ & 924.30 & 4.02$\times$ & 0.9830 \textcolor{black!40}{\scriptsize ($-$1.0\%)} & 32.25 \textcolor{black!40}{\scriptsize ($-$1.1\%)} \\
        $\textbf{\texttt{DuCa}}$ $(\mathcal{N}=5)$~\cite{zou2024DuCa} & 7.32 & 3.15$\times$ & 978.76 & 3.80$\times$ & 0.9982 \textcolor{black!40}{\scriptsize ($+$0.5\%)} & 32.41 \textcolor{black!40}{\scriptsize ($-$0.6\%)} \\
        \textbf{TeaCache} $({l}=0.8)$~\cite{liu2024timestep} & 6.42 & 3.58$\times$ & \underline{892.35} & \underline{4.17}$\times$ & 0.8710 \textcolor{black!40}{\scriptsize ($-$12.3\%)} & 31.89 \textcolor{black!40}{\scriptsize ($-$2.2\%)} \\
        $\textbf{DBCache}$ $(\mathcal{F}=4,B=4)$~\cite{cache-dit@2025} & \underline{5.92} & \underline{3.90}$\times$ & 907.20 & 4.10$\times$ & 0.6372 \textcolor{black!40}{\scriptsize ($-$35.8\%)} & 32.10 \textcolor{black!40}{\scriptsize ($-$1.6\%)} \\
        $\textbf{TaylorSeer}$ $(\mathcal{N}=4, O=2)$~\cite{liu2025reusingforecastingacceleratingdiffusion} & 8.31 & 2.80$\times$ & 967.91 & 3.84$\times$ & 0.9887 \textcolor{black!40}{\scriptsize ($-$0.4\%)} & 32.58 \textcolor{black!40}{\scriptsize ($-$0.1\%)} \\
        $\textbf{FoCa}$ $(\mathcal{N}=4)$~\cite{zhengFoCa2025} & 8.37 & 2.76$\times$ & 1050.70 & 3.54$\times$ & 0.9782 \textcolor{black!40}{\scriptsize ($-$1.5\%)} & 32.73 \textcolor{black!40}{\scriptsize ($+$0.4\%)} \\
        $\textbf{Clusca}$ $(\mathcal{N}=4, O=2, K=16)$~\cite{Zheng2025Compute} & 8.27 & 2.79$\times$ & 1045.58 & 3.56$\times$ & 0.9876 \textcolor{black!40}{\scriptsize ($-$0.5\%)} & 32.62 \textcolor{black!40}{\scriptsize ($+$0.0\%)} \\
        $\textbf{HyCa}$ $(\mathcal{N}=5)$~\cite{zheng2025let} & 6.83 & 3.38$\times$ & 893.54 & 4.16$\times$ & \underline{1.0096} \textcolor{black!40}{\scriptsize ($+$1.7\%)} & \underline{32.87} \textcolor{black!40}{\scriptsize ($+$0.8\%)} \\
        \rowcolor{gray!100}
        $\textbf{LinCa}$ $(\mathcal{N}=6)$ & \bf 5.27 & \textbf{4.38$\times$} & \bf 823.21 & \textbf{4.52$\times$} &\bf 1.0228 \textcolor[HTML]{0f98b0}{\scriptsize ($+$3.0\%)} & \bf 32.97 \textcolor[HTML]{0f98b0}{\scriptsize ($+$1.1\%)} \\
        \midrule

        $\textbf{FORA}$ $(\mathcal{N}=5)$~\cite{selvaraju2024fora} & 6.94 & 3.33$\times$ & 893.54 & 4.16$\times$ & 0.8276 \textcolor{black!40}{\scriptsize ($-$16.7\%)} & 31.95 \textcolor{black!40}{\scriptsize ($-$2.0\%)} \\
        $\textbf{\texttt{ToCa}}$ $(\mathcal{N}=8)$~\cite{zou2024accelerating} &  10.17 & 2.27$\times$ & 784.54 & 4.74$\times$ & 0.9479 \textcolor{black!40}{\scriptsize ($-$4.5\%)} & 32.16 \textcolor{black!40}{\scriptsize ($-$1.4\%)} \\
        $\textbf{\texttt{DuCa}}$ $(\mathcal{N}=7)$~\cite{zou2024DuCa} & 6.06 & 3.83$\times$ & 760.14 & 4.89$\times$ & 0.9785 \textcolor{black!40}{\scriptsize ($-$1.5\%)} & 32.26 \textcolor{black!40}{\scriptsize ($-$1.1\%)} \\
        \textbf{TeaCache} $({l}=1)$~\cite{liu2024timestep} & 7.24 & 3.19$\times$ & \underline{743.63} & \underline{5.01}$\times$ & 0.8421 \textcolor{black!40}{\scriptsize ($-$15.2\%)} & 32.01 \textcolor{black!40}{\scriptsize ($-$1.8\%)} \\
        $\textbf{DBCache}$ $(\mathcal{F}=4,B=2)$~\cite{cache-dit@2025} & \underline{5.25} & \underline{4.40}$\times$ & 793.07 & 4.69$\times$ & 0.5106 \textcolor{black!40}{\scriptsize ($-$48.6\%)} & 32.01 \textcolor{black!40}{\scriptsize ($-$1.8\%)} \\
        $\textbf{TaylorSeer}$ $(\mathcal{N}=5, O=2)$~\cite{liu2025reusingforecastingacceleratingdiffusion} & 6.71 & 3.46$\times$ & 893.54 & 4.16$\times$ & 0.9793 \textcolor{black!40}{\scriptsize ($-$1.4\%)} & 32.63 \textcolor{black!40}{\scriptsize ($+$0.1\%)} \\
        $\textbf{FoCa}$ $(\mathcal{N}=6)$~\cite{zhengFoCa2025} & 6.76 & 3.42$\times$ & 745.39 & 4.99$\times$ & 0.9741 \textcolor{black!40}{\scriptsize ($-$1.9\%)} & \textbf{33.10} \textcolor{black!40}{\scriptsize ($+$1.5\%)} \\
        $\textbf{Speca}$ $(\mathcal{N}_{\text{max}}=8, \mathcal{N}_{\text{min}}=2)$~\cite{Liu2025SpeCa} & 6.61 & 3.48$\times$ & 791.38 & 4.70$\times$ & 1.0012 \textcolor{black!40}{\scriptsize ($+$0.8\%)} & 32.45 \textcolor{black!40}{\scriptsize ($-$0.5\%)} \\
        $\textbf{Clusca}$ $(\mathcal{N}=5, O=1, K=16)$~\cite{Zheng2025Compute} & 6.31 & 3.66$\times$ & 897.03 & 4.14$\times$ & 0.9748 \textcolor{black!40}{\scriptsize ($-$1.8\%)} & 32.50 \textcolor{black!40}{\scriptsize ($-$0.3\%)} \\
        $\textbf{HyCa}$ $(\mathcal{N}=6)$~\cite{zheng2025let} & 6.09 & 3.79$\times$ & 744.81 & 5.00$\times$ & \underline{1.0043} \textcolor{black!40}{\scriptsize ($+$1.1\%)} & 32.64 \textcolor{black!40}{\scriptsize ($+$0.1\%)} \\
        \rowcolor{gray!100}
        $\textbf{LinCa}$ $(\mathcal{N}=8)$ & \bf 4.40 & \textbf{5.25$\times$} & \bf 674.64 & \textbf{5.51$\times$} &\bf 1.0162 \textcolor[HTML]{0f98b0}{\scriptsize ($+$2.3\%)} & \underline{32.72} \textcolor[HTML]{0f98b0}{\scriptsize ($+$0.3\%)} \\
        
\bottomrule

\end{tabular}}

\vspace{-3mm}

\label{table:FLUX}
\footnotesize
\end{table*}

\begin{figure}[htbp]
    \centering
    \includegraphics[width=0.9\linewidth]
    {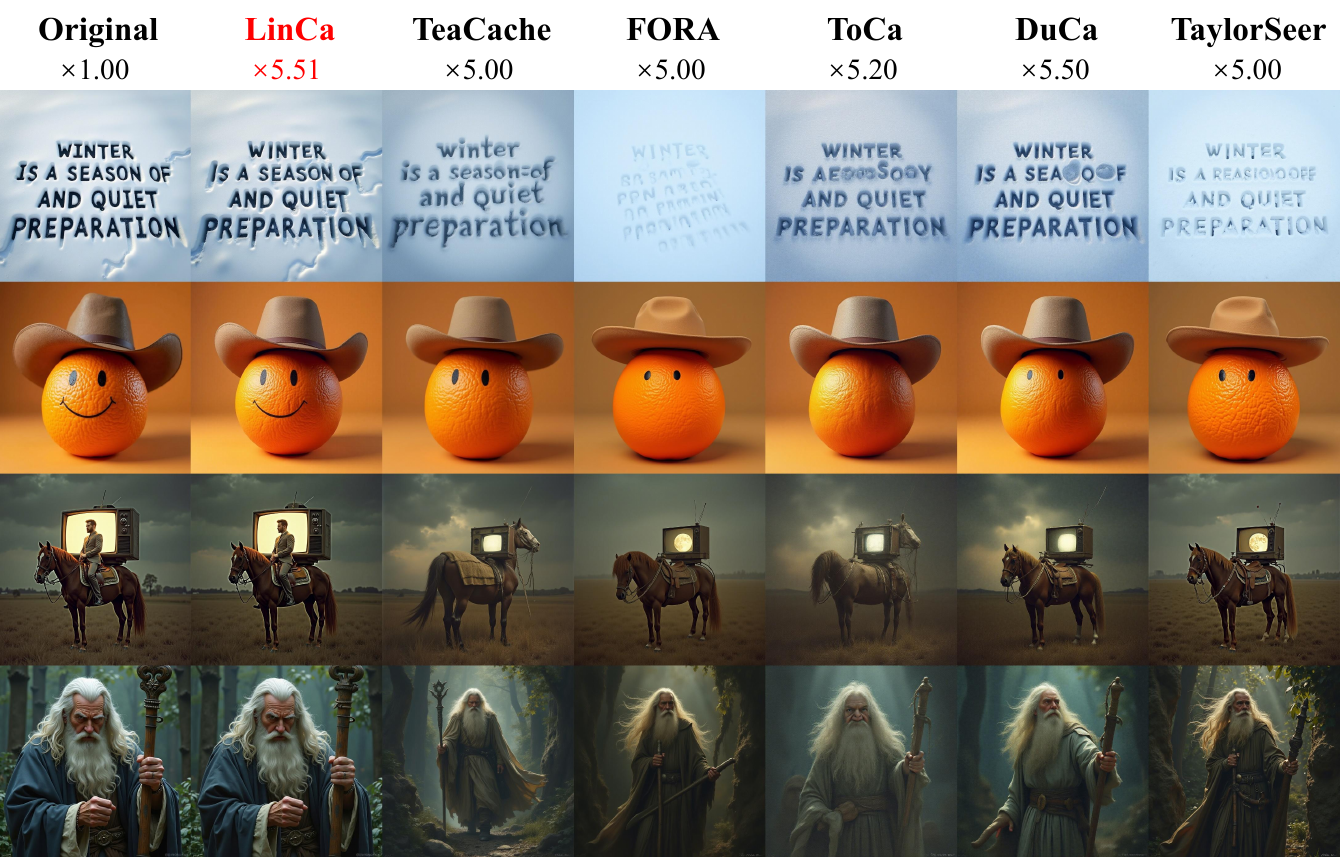}
    \captionof{figure}{On FLUX.1-dev, LinCa delivers higher speedup with better text consistency, fine-grained details, and spatial relationships.}

    \label{fig:flux-dev}
\end{figure}

 \vspace{4mm}

\begin{table*}[h!]
\centering
\caption{\textbf{Quantitative comparison of text-to-image generation} for Qwen-Image. Best results are highlighted in \textbf{bold}, and second-best are \underline{underlined}.}
\vspace{-3mm}
  \resizebox{\textwidth}{!}{ 
    \begin{tabular}{l | c  c | c  c | c  c | c c c}
        \toprule
        \multirow[c]{2}{*}{\bf Method}
        & \multicolumn{4}{c|}{\textbf{Acceleration}} 
        & \multicolumn{2}{c|}{\textbf{Quality Metrics}} 
        & \multicolumn{3}{c}{\textbf{Perceptual Metrics}}\rule{0pt}{2ex}\\
        \cline{2-10}
        & \textbf{Latency(s) \(\downarrow\)} 
        & \textbf{Speed \(\uparrow\)} 
        & \textbf{FLOPs(T) \(\downarrow\)}  
        & \textbf{Speed \(\uparrow\)} 
        & \textbf{ImageReward\(\uparrow\)} 
        & \textbf{CLIP\(\uparrow\)} 
        & \textbf{PSNR\(\uparrow\)} 
        & \textbf{SSIM\(\uparrow\)}
        & \textbf{LPIPS\(\downarrow\)}\rule{0pt}{2ex}\\
        \midrule

\textbf{Original: 50 steps}
& 126.60 & 1.00$\times$ & 12917.56 & 1.00$\times$ & 1.2532 \textcolor{black!40}{\scriptsize ($+$0.0\%)} & 35.52 & $\infty$ & 1.00 & 0.00 \\

\textbf{$50\%$ steps }
& 63.62 & 1.99$\times$ & 6458.78 & 2.00$\times$ & 1.2023 \textcolor{black!40}{\scriptsize ($-$4.1\%)} & 35.24 & 30.55 & 0.75 & 0.27 \\

\textbf{$20\%$ steps} 
& 25.89 & 4.89$\times$ & 2583.51 & 5.00$\times$ & 0.9223 \textcolor{black!40}{\scriptsize ($-$26.4\%)} & 34.94 & 28.60 & 0.60 & 0.52 \\
\midrule

\textbf{TaylorSeer }($\mathcal{N}$=3)~\cite{liu2025reusingforecastingacceleratingdiffusion}
& 62.36 & 2.03$\times$ & \textbf{4646.60} & \textbf{2.78}$\times$ & 1.0674 \textcolor{black!40}{\scriptsize ($-$14.8\%)} & 34.78 & 28.14 & 0.53 & 0.65 \\

\textbf{HyCa }($\mathcal{N}$=3)~\cite{zheng2025let}
& \underline{59.72} & \underline{2.12}$\times$ & \textbf{4646.60} & \textbf{2.78}$\times$ & \underline{1.2316} \textcolor{black!40}{\scriptsize ($-$1.7\%)} & \underline{35.01} & \underline{30.21} & \underline{0.80} & \underline{0.25} \\

\rowcolor{gray!100}
\textbf{LinCa }($\mathcal{N}$=3)
& \bf {58.88} & \bf {2.15}$\times$ & \underline{4705.76} & \underline{2.75}$\times$ & \bf {1.2329} \textcolor[HTML]{0f98b0}{\scriptsize ($-$1.6\%)} & \bf {35.40} & \bf {31.86} & \bf {0.83} & \bf {0.18} \\
\midrule

\textbf{FORA }($\mathcal{N}$=4)~\cite{selvaraju2024fora}
& 38.13 & 3.32$\times$ & 3359.99 & 3.84$\times$ & 0.9315 \textcolor{black!40}{\scriptsize ($-$25.7\%)} & 34.33 & 28.65 & 0.60 & 0.50 \\

\textbf{\texttt{ToCa} }($\mathcal{N}$=8, $\mathcal{R}$=75\%)~\cite{zou2024accelerating}
& 60.87 & 2.08$\times$ & 2991.34 & 4.32$\times$ & 1.0218 \textcolor{black!40}{\scriptsize ($-$18.5\%)} & \underline{34.97} & 28.94 & 0.63 & 0.45 \\

\textbf{\texttt{DuCa} }($\mathcal{N}$=9, $\mathcal{R}$=80\%)~\cite{zou2024DuCa}
& 34.50 & 3.67$\times$ & 2958.13 & 4.37$\times$ & 0.7709 \textcolor{black!40}{\scriptsize ($-$38.5\%)} & 34.59 & 28.44 & 0.58 & 0.54 \\

\textbf{TaylorSeer }($\mathcal{N}$=6)~\cite{liu2025reusingforecastingacceleratingdiffusion}
& \underline{30.58} & \underline{4.14}$\times$ & \textbf{2583.97} & \textbf{5.00}$\times$ & 1.0125 \textcolor{black!40}{\scriptsize ($-$19.2\%)} & 34.77 & 28.59 & 0.62 & 0.45 \\

\textbf{HyCa }($\mathcal{N}$=6)~\cite{zheng2025let}
& 36.48 & 3.47$\times$ & \underline{2584.46} & \underline{5.00}$\times$ & \underline{1.1967} \textcolor{black!40}{\scriptsize ($-$4.5\%)} & 34.89 & \underline{29.68} & \underline{0.71} & \underline{0.31} \\


\rowcolor{gray!100}
\textbf{LinCa }($\mathcal{N}$=6)
& \bf {30.51} & \bf {4.15}$\times$ & 2635.31 & 4.90$\times$ & \bf {1.2163} \textcolor[HTML]{0f98b0}{\scriptsize ($-$2.9\%)} & \bf {35.34} & \bf {29.84} & \bf {0.73} & \bf {0.29} \\

\midrule

\textbf{FORA }($\mathcal{N}$=6)~\cite{selvaraju2024fora}
& 28.51 & 4.44$\times$ & 2326.74 & 5.55$\times$ & 0.4812 \textcolor{black!40}{\scriptsize ($-$61.6\%)} & 33.31 & 28.48 & 0.55 & 0.59 \\

\textbf{\texttt{ToCa} }($\mathcal{N}$=12, $\mathcal{R}$=85\%)~\cite{zou2024accelerating}
& 50.64 & 2.50$\times$ & 2406.20 & 5.37$\times$ & 0.5511 \textcolor{black!40}{\scriptsize ($-$56.0\%)} & 34.05 & 28.68 & 0.58 & 0.54 \\

\textbf{\texttt{DuCa} }($\mathcal{N}$=12, $\mathcal{R}$=90\%)~\cite{zou2024DuCa}
& 28.39 & 4.46$\times$ & 2171.56 & 5.95$\times$ & 0.4104 \textcolor{black!40}{\scriptsize ($-$67.3\%)} & 33.34 & 28.38 & 0.57 & 0.61 \\

\textbf{TaylorSeer }($\mathcal{N}$=9)~\cite{liu2025reusingforecastingacceleratingdiffusion}
& 24.49 & 5.17$\times$ & 2067.29 & 6.25$\times$ & 0.7318 \textcolor{black!40}{\scriptsize ($-$41.6\%)} & 32.89 & 28.24 & 0.56 & 0.57 \\

\textbf{HyCa }($\mathcal{N}$=8)~\cite{zheng2025let}
& \underline{23.53} & \underline{5.38}$\times$ & \underline{2066.82} & \underline{6.25}$\times$ & \underline{1.0432} \textcolor{black!40}{\scriptsize ($-$16.8\%)} & \underline{34.82} & \underline{28.86} & \underline{0.62} & \underline{0.44} \\

\rowcolor{gray!100}
\textbf{LinCa }($\mathcal{N}$=10)
& \bf {23.19} & \bf {5.46}$\times$ & \bf {1859.35} & \bf {6.95}$\times$ & \bf {1.0524} \textcolor[HTML]{0f98b0}{\scriptsize ($-$16.0\%)} & \bf {35.17} & \bf {29.10} & \bf {0.69} & \bf {0.39} \\

\midrule

\end{tabular}
}
\label{table:qwen-image-Metrics}

\end{table*}


\begin{figure}[!t]
    \centering
    \includegraphics[width=0.9\linewidth]
    {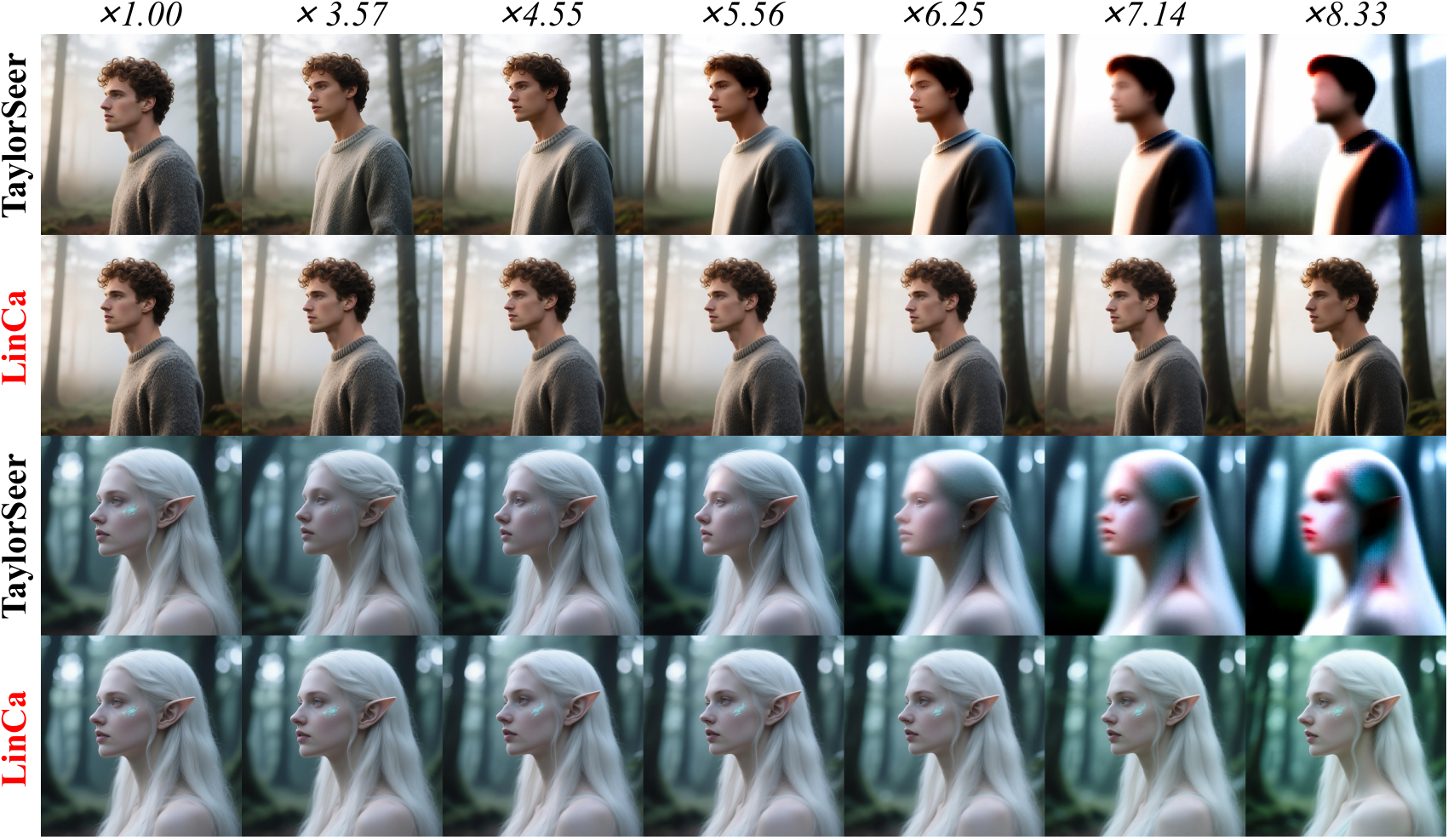}
    \captionof{figure}{On Qwen-Image, LinCa retains superior image quality and details as the acceleration ratio increases, while Taylorseer degrades with blurring and detail loss.}

    \label{fig:qwen}
    \vspace{-3mm}
\end{figure}

As shown in Table~\ref{table:FLUX}, LinCa achieves the best speed-quality trade-off on \textbf{FLUX.1-dev}. At $N=4$, it reaches an ImageReward of $\textbf{1.0175}$ with $\textbf{3.32}\times$ acceleration, surpassing FoCa ($0.9917$ at $2.80\times$) and TaylorSeer ($1.0018$ at $2.82\times$). At $N=6$, LinCa demonstrates remarkable robustness, achieving $\textbf{1.0228}$ at $\textbf{4.52}\times$, which outperforms both HyCa ($1.0096$ at $4.16\times$) and Clusca ($0.9876$ at $3.56\times$). At $N=8$, it maintains a superior quality of $\textbf{1.0162}$ at $\textbf{5.51}\times$. Other methods such as DBCache and TeaCache suffer from substantial degradation. Visual comparison in Fig.~\ref{fig:flux-dev} further confirms LinCa's advantage in preserving image details and text-alignment under high compression ratios.

In Table~\ref{table:qwen-image-Metrics}, LinCa consistently achieves the best overall trade-off on \textbf{Qwen-Image} across all acceleration levels. At $N=3$, it is comparable to TaylorSeer in speed ($\textbf{2.75}\times$) but yields higher quality (ImageReward $\textbf{1.2329}$ vs. $1.0674$, and highest PSNR $\textbf{31.86}$). At $N=6$, LinCa remains strong ($\textbf{1.2163}$, $\textbf{29.84}$), outperforming HyCa ($1.1967$, $29.68$) and surpassing other methods like ToCa ($1.0218$) and FORA ($0.9315$). At $N=10$, it sustains high quality ($\textbf{1.0524}$ at $\textbf{6.95}\times$), while others drop sharply. These results highlight LinCa's robustness under high acceleration while preserving superior visual fidelity and text-alignment. Visual comparison in Fig.~\ref{fig:qwen} further demonstrates that LinCa consistently preserves high image quality and intricate details even as the acceleration ratio increases.

\subsection{Results on Text-to-Video Generation}

\vspace{-3mm}


\begin{table*}[htbp]
\centering

\vspace{-3mm}
\caption{\textbf{Quantitative comparison of text-to-video generation} for HunyuanVideo. Best results are highlighted in \textbf{bold}, and second-best are \underline{underlined}.}
\vspace{-2mm}
\resizebox{\textwidth}{!}{
\begin{tabular}{l | c | c  c | c  c | c }
    \toprule
    \multirow[c]{2}{*}{\bf Method} & {\bf Efficient} &\multicolumn{4}{c|}{\bf Acceleration} &{\bf VBench $\uparrow$}  \\
    \cline{3-6}
    & {\bf Attention } & {\bf Latency(s) $\downarrow$} & {\bf Speed $\uparrow$} & {\bf FLOPs(T) $\downarrow$}  & {\bf Speed $\uparrow$} & \bf Score(\%)\rule{0pt}{2ex}\\ 
    \midrule

$\textbf{Original: 50 steps}$ 
                           & \ding{52}  & 185.00 & 1.00$\times$& {29773.0}   & {1.00$\times$} & 80.66 \textcolor{black!40}{\scriptsize ($+$0.0\%)}      \\

{$22\%$\textbf{ steps}}  & \ding{52}  & 40.66 & 4.55$\times$ & {6550.1}   & {4.55$\times$} & 78.74 \textcolor{black!40}{\scriptsize ($-$2.4\%)}          \\

\midrule


$\textbf{FORA}$ $(\mathcal{N}=5)$~\cite{selvaraju2024fora}
                            & \ding{52}  & 43.84 & 4.22$\times$  & {5960.4}   & {5.00$\times$} & 78.83 \textcolor{black!40}{\scriptsize ($-$2.3\%)}     \\

$\textbf{\texttt{ToCa}}$ $(\mathcal{N}=5,R=90\%)$~\cite{zou2024accelerating}
                           & \ding{56}  & 49.07 & 3.77$\times$ & {7006.2}   & {4.25$\times$} & 78.86 \textcolor{black!40}{\scriptsize ($-$2.2\%)}    \\
  
$\textbf{\texttt{DuCa}} $ $(\mathcal{N}=5,R=90\%)$~\cite{zou2024DuCa} 
                           & \ding{52}  & 40.39 & 4.58$\times$ & {6483.2}   & {4.48$\times$} & 78.72 \textcolor{black!40}{\scriptsize ($-$2.4\%)}    \\
\textbf{TeaCache} $({l}=0.4)$~\cite{liu2024timestep} 
                           & \ding{52} & \underline{38.87} & \underline{4.76}$\times$ & 6550.1 & 4.55$\times$ & 79.36 \textcolor{black!40}{\scriptsize ($-$1.6\%)}  \\
$\textbf{TaylorSeer} $ $(\mathcal{N}=5,O=1)$~\cite{liu2025reusingforecastingacceleratingdiffusion} 
                           & \ding{52}  & 44.47 & 4.16$\times$ & {5960.4}  & {5.00$\times$} & 79.93 \textcolor{black!40}{\scriptsize ($-$0.9\%)} \\

$\textbf{Speca} $ $(\mathcal{N}_{\text{max}}=8, \mathcal{N}_{\text{min}}=2)$~\cite{Liu2025SpeCa} 
                           & \ding{52}  & 44.05 & 4.20$\times$ & {\underline{5692.7}}  & {\underline{5.23}$\times$} & 79.98 \textcolor{black!40}{\scriptsize ($-$0.8\%)} \\

$\textbf{Clusca} $ $(\mathcal{N}=5,K=32)$~\cite{Zheng2025Compute} 
                           & \ding{52}  & 48.30 & 3.83$\times$ & {5968.1}  & {4.99$\times$} & \underline{79.99} \textcolor{black!40}{\scriptsize ($-$0.8\%)} \\

$\textbf{FoCa} $ $(\mathcal{N}=5)$~\cite{zhengFoCa2025} 
                           & \ding{52}  & 44.05 & 4.20$\times$ & {5966.5}  & {4.99$\times$} & 79.96 \textcolor{black!40}{\scriptsize ($-$0.9\%)} \\

\rowcolor{gray!100}
$\textbf{LinCa} $ $(\mathcal{N}=6)$ 
                           & \ding{52}  & \bf 38.21 & \textbf{4.84$\times$} & \bf 5413.27  & \textbf{5.50$\times$} & \bf {80.16} \textcolor[HTML]{0f98b0}{\scriptsize ($-$0.6\%)}  \\

    \bottomrule
\end{tabular}}
\label{table:HunyuanVideo-Metrics}
\end{table*}

\vspace{-6mm}

\begin{figure}[htbp]
    \centering
     \vspace{-2mm}
    \includegraphics[width=0.9\linewidth]
    {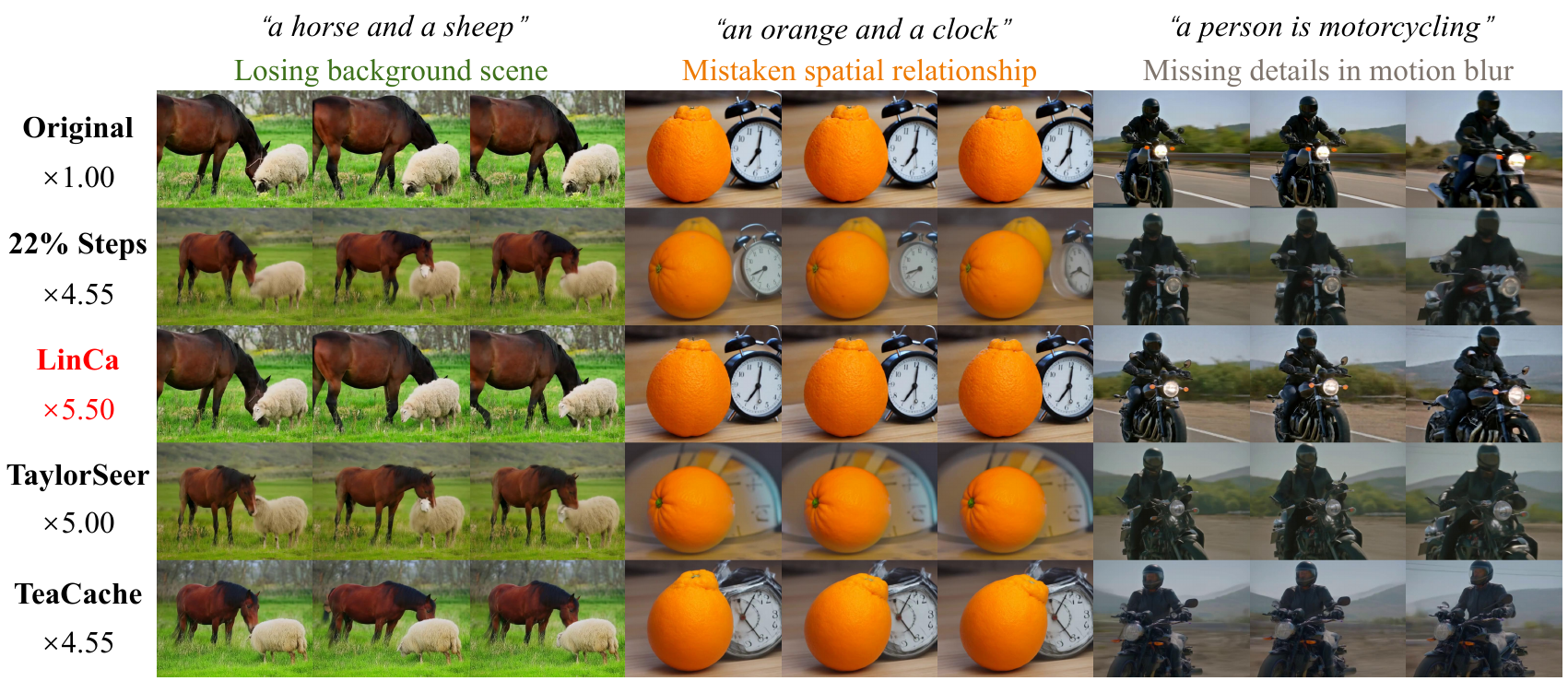}
    \captionof{figure}{On HunyuanVideo, LinCa maintains high-quality generation under higher acceleration ratio, while other methods suffer from lossing background scene, mistaken spatial relationships, and missing motional details.}

    \label{fig:hunyuan}
    \vspace{-2mm}
\end{figure}

As shown in Table~\ref{table:HunyuanVideo-Metrics}, LinCa exhibits superior quality on \textbf{HunyuanVideo}. At $\mathcal{N}=6$, it achieves a notable $\textbf{5.50}\times$ speedup while preserving a competitive VBench score of \textbf{80.16}, representing only a slight decrease from the 50-step baseline (80.66). In contrast, Clusca and Speca reach 79.99 and 79.98 with less speedup, while TaylorSeer reaches only 5.00$\times$ with 79.93 and ToCa/DuCa degrade further. This demonstrates an optimal balance between inference speed and visual fidelity in high-quality video generation. Qualitative comparisons on Fig.~\ref{fig:hunyuan} further confirm that LinCa preserves video quality, background scenery, spatial relationship and motional details better.


\subsection{Results on Image Editing}


\begin{table*}[!htbp]
\centering
\caption{\textbf{Quantitative comparison of image editing} for Qwen-Image-Edit. Best results are highlighted
in \textbf{bold}, and second-best results are \underline{underlined}.}
\vspace{-3mm}
  \resizebox{\textwidth}{!}{ 
    \begin{tabular}{l | c  c | c  c | c c c|c c c}
    \toprule
    \multirow[c]{2}{*}{\bf Method}
    & \multicolumn{4}{c|}{\bf Acceleration} 
    & \multicolumn{3}{c|}{\bf GEdit-CN (Full)}
    & \multicolumn{3}{c}{\bf GEdit-EN (Full)} \rule{0pt}{2ex}\\
    \cline{2-11}
    & {\bf Latency(s) $\downarrow$} 
    & {\bf Speed $\uparrow$} 
    & {\bf FLOPs(T) $\downarrow$}  
    & {\bf Speed $\uparrow$} 
    & {\bf SC $\uparrow$} 
    & {\bf PQ $\uparrow$} 
    & {\bf OS $\uparrow$}
    & {\bf SC $\uparrow$} 
    & {\bf PQ $\uparrow$} 
    & {\bf OS $\uparrow$}\rule{0pt}{2ex}\\
    \midrule

\textbf{Original: 50 steps}
& 284.51  & 1.00$\times$ & 28190.88 & 1.00$\times$ & 7.68 & 7.51 & 7.41  & 7.82 & 7.54 & 7.54 \\

50\% \textbf{steps} 
& 143.29  & 1.99$\times$ & 14095.44 & 2.00$\times$ & 7.70 & 7.53 & 7.44  & 7.77 & 7.52 & 7.47 \\

20\% \textbf{steps} 
& 58.45  & 4.87$\times$ & 5638.18 & 5.00$\times$ & 7.65 & 7.42 & 7.35  & 7.73 & 7.46 & 7.44 \\

\midrule
\textbf{FORA }($\mathcal{N}$=5)~\cite{selvaraju2024fora}
& \underline{63.15}  & \underline{4.51}$\times$ & \underline{5643.13} & \underline{5.00}$\times$ & 7.60 & 7.31 & 7.25 & 7.62 & 7.34 & 7.28 \\

\textbf{\texttt{DuCa} }($\mathcal{N}$=7, $\mathcal{R}$=95\%)~\cite{zou2024DuCa} & 69.54  & 4.09$\times$ & 5699.89 & 4.95$\times$ & \textbf{7.73} & \underline{7.44} & \underline{7.44}  & \underline{7.80} & \underline{7.40} & \underline{7.45}  \\

\textbf{TaylorSeer }($\mathcal{N}$=6)~\cite{liu2025reusingforecastingacceleratingdiffusion}
&  65.66  & 4.33$\times$ & \underline{5643.13} & \underline{5.00}$\times$ & 7.25 & 7.09 & 6.92  & 7.26 & 7.14 & 6.89 \\

\rowcolor{gray!100}
\textbf{LinCa }($\mathcal{N}$=7)
& \textbf{60.02}  & \textbf{4.74}$\times$ & \textbf{5110.58} & \textbf{5.52}$\times$ & \underline{7.63} & \textbf{7.52} & \textbf{7.45}  & \textbf{7.81} & \textbf{7.55} & \textbf{7.56} \\

\midrule

\textbf{FORA }($\mathcal{N}$=7)~\cite{selvaraju2024fora}
& \underline{52.20}  & \underline{5.45}$\times$ & \underline{4515.74} & \underline{6.24}$\times$ & 7.42 & \underline{7.13} & \underline{7.06}  & 7.43 & \underline{7.19} & \underline{7.06} \\

\textbf{\texttt{DuCa} }($\mathcal{N}$=10, $\mathcal{R}$=95\%)~\cite{zou2024DuCa} & 59.81  & 4.76$\times$ & 5158.45 & 5.46$\times$ & \underline{7.50} & 5.75 & 6.39  & \underline{7.52} & 5.77 & 6.41  \\

\textbf{TaylorSeer }($\mathcal{N}$=9)~\cite{liu2025reusingforecastingacceleratingdiffusion}
& 53.92  & 5.28$\times$ & \underline{4515.74}& \underline{6.24}$\times$ & 6.61 & 6.65 & 6.31  & 6.67 & 6.63 & 6.31 \\

\rowcolor{gray!100}
\textbf{LinCa }($\mathcal{N}$=10)
& \textbf{49.05}  & \textbf{5.80}$\times$ & \textbf{3982.82} & \textbf{7.08}$\times$& \textbf{7.55} & \textbf{7.40} & \textbf{7.27}  & \textbf{7.68} & \textbf{7.46} & \textbf{7.40} \\

\bottomrule
\end{tabular}
}

\label{table:qwen-image-edit-Metrics}

\end{table*}

\vspace{-2mm}

\begin{figure}[!htbp]
    \centering
 \vspace{-8mm}
    \includegraphics[width=0.9\linewidth]
    {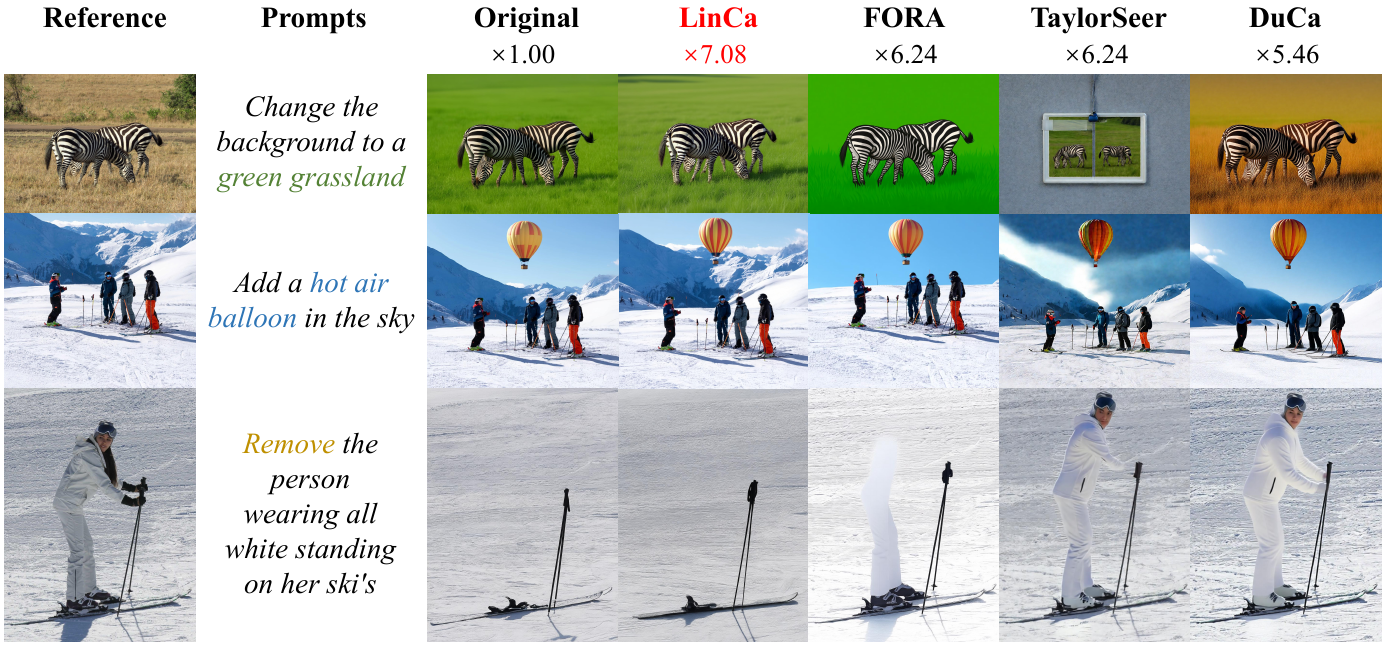}
    \captionof{figure}{On Qwen-Image-Edit, LinCa delivers better prompt comprehension and generates high-fidelity images while maintaining consistency in non-edited regions.}

    \label{fig:qwen-edit}
    \vspace{-2mm}
\end{figure}

\FloatBarrier

As shown in Table~\ref{table:qwen-image-edit-Metrics}, LinCa exhibits exceptional performance on \textbf{Qwen-Image-Edit}. At $N=7$, it obtains overall scores of \textbf{7.45} (CN) and \textbf{7.56} (EN), outperforming TaylorSeer (6.92/6.89), FORA (7.25/7.28), and DuCa (7.44/7.45), even exceeding the original model's performance. At $N=10$, LinCa continues to lead with \textbf{7.27/7.40}, whereas other baselines such as DuCa and TaylorSeer drop to 6.39/6.41 and 6.31/6.31. These results demonstrate that LinCa remains robust under high acceleration, effectively preserving high-fidelity synthesis. Qualitative comparisons on Fig.~\ref{fig:qwen-edit} further confirm its effectiveness in handling different editing tasks with high quality and consistency in non-edited regions.

\subsection{Results on Distilled or Quantized Models}



\begin{table*}[!htbp]
\centering
\vspace{-3mm}
\caption{\textbf{Quantitative comparison of text-to-image generation} for FLUX.1-lite-8B, FLUX.1-schnell and FLUX.1-dev-int8.}
\vspace{-3mm}
  \resizebox{\textwidth}{!}{ 
    \begin{tabular}{l | c  c | c  c | c | c}
        \toprule
        \multirow{2}{*}{\textbf{Method}}
        & \multicolumn{4}{c|}{\textbf{Acceleration}} 
        & \multirow{2}{*}{\textbf{ImageReward\(\uparrow\)}} 
        & \multirow{2}{*}{\textbf{CLIP Score\(\uparrow\)}}\rule{0pt}{2ex}\\
        \cline{2-5}
        & \textbf{Latency(s) \(\downarrow\)} 
        & \textbf{Speed \(\uparrow\)} 
        & \textbf{FLOPs(T) \(\downarrow\)}  
        & \textbf{Speed \(\uparrow\)} 
        & & \rule{0pt}{2ex}\\
        \midrule

\textbf{FLUX.1-lite-8B: 28 steps} 
& 8.21 & 1.00$\times$ & 1291.49 & 1.00$\times$ & 0.8936 \textcolor{black!40}{\scriptsize (+0.0\%)} & 32.12 \textcolor{black!40}{\scriptsize (+0.0\%)} \\

\rowcolor{gray!100}
\textbf{LinCa ($\mathcal{N}$=3): 28 steps}& \bf 3.34 & \bf 2.46$\times$ & \bf 556.21 & \bf 2.32$\times$ & \bf 0.9070 \textcolor[HTML]{0f98b0}{\scriptsize (+1.5\%)} & \bf 32.36 \textcolor[HTML]{0f98b0}{\scriptsize (+0.7\%)} \\

\midrule

\textbf{FLUX.1-schnell: 4 steps} 
& 2.48 & 1.00$\times$ & 283.56 & 1.00$\times$ & 0.9692 \textcolor{black!40}{\scriptsize (+0.0\%)} & 32.54 \textcolor{black!40}{\scriptsize (+0.0\%)} \\

\rowcolor{gray!100}
\textbf{LinCa ($\mathcal{N}$=3): 4 steps}
& \bf 1.38 & \bf 1.80$\times$ & \bf 142.16 & \bf 1.99$\times$ & \bf 0.9843 \textcolor[HTML]{0f98b0}{\scriptsize (+1.6\%)} & \bf 32.67 \textcolor[HTML]{0f98b0}{\scriptsize (+0.4\%)} \\

\midrule

\textbf{FLUX.1-dev-int8: 50 steps} 
& 12.55 & 1.00$\times$ & 1888.07 & 1.00$\times$ & 0.9744 \textcolor{black!40}{\scriptsize (+0.0\%)} & 32.55 \textcolor{black!40}{\scriptsize(+0.0\%)} \\

\rowcolor{gray!100}
\textbf{LinCa ($\mathcal{N}$=3): 50 steps}&  \bf 4.61 & \bf 2.72$\times$ & \bf 718.17 & \bf 2.63$\times$ & \bf 1.0036 \textcolor[HTML]{0f98b0}{\scriptsize (+3.0\%)} & \bf 32.81 \textcolor[HTML]{0f98b0}{\scriptsize(+0.8\%)} \\

\bottomrule

\end{tabular}
}

\label{table:distill}
\raggedright

\vspace{-2mm}
\end{table*}

\FloatBarrier

To validate the generality of LinCa, we evaluate its performance when integrated with other mainstream acceleration techniques, specifically model distillation (\textbf{FLUX.1-lite-8B}), step distillation (\textbf{FLUX.1-schnell}), and quantization (\textbf{FLUX.1-dev-int8}). As shown in Table~\ref{table:distill}, LinCa consistently improves performance across all settings without degrading generation quality. For the model-distilled variant, LinCa achieves a \textbf{2.32}$\times$ speed-up, increasing the ImageReward to \textbf{0.9070} and the CLIP Score to \textbf{32.36}. When applied to the step-distilled model, our framework delivers a \textbf{1.99}$\times$ acceleration, reaching an \textbf{0.9843} ImageReward and \textbf{32.67} CLIP Score. Furthermore, integrating LinCa with INT8 quantization provides a \textbf{2.63}$\times$ speed improvement while maintaining high fidelity (\textbf{1.0036} ImageReward, \textbf{32.81} CLIP Score). These results demonstrate the broad compatibility of LinCa with various mainstream acceleration methods, serving as an effective complementary module that further enhances acceleration performance while maintaining high generation quality.

\subsection{Ablation Study}


\begin{figure}[htbp]
    \centering
    \includegraphics[width=0.9\linewidth]
    {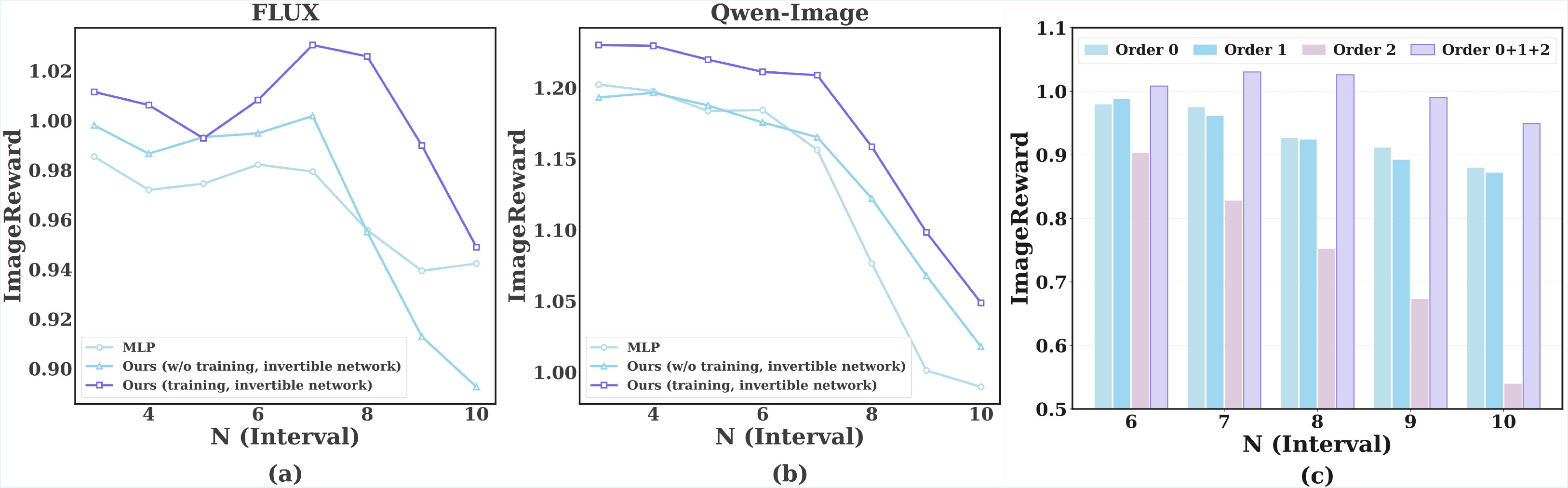}

    \captionof{figure}{\textbf{Ablation results of LinCa. (a)-(b)}: Comparison of decomposition network architectures on FLUX.1-dev and Qwen-Image. The learnable invertible network consistently outperforms both MLP and untrained alternatives. \textbf{(c)} Comparison of prediction order strategies. Differentiated multi-order prediction maintains superior quality over single-order prediction.}

    \label{ablation}

\end{figure}

\textbf{Decomposition Network Ablation.} We study the impact of different decomposition networks on generation quality on FLUX.1-dev and Qwen-Image. As shown in Figure~\ref{ablation}(a)-(b), LinCa's learnable invertible network architecture achieves the best results across all intervals. Compared to ordinary non-invertible networks such as learnable MLPs, the invertible network yields higher overall quality, indicating that its lossless reconstruction property effectively avoids information loss during the mapping process. Compared to non-learnable strategies that exhibit significant quality degradation under high acceleration ratios, the learnable approach demonstrates more stable performance, adapting to the specific feature dynamics.

\textbf{Prediction Strategy Ablation.} We further compare using $0^{th}$ order (pure reuse), $1^{st}$ order, and $2^{nd}$ order prediction individually against LinCa's combined $0^{th}+1^{st}+2^{nd}$ order prediction, to investigate the impact of different prediction strategies on generation quality. As shown in Figure~\ref{ablation}(c), LinCa's differentiated order prediction strategy outperforms single-order configurations across all intervals. Notably, as $N$ increases, uniform single-order prediction, especially $2^{nd}$ order prediction alone, suffers from significant quality degradation, while LinCa accurately separates features into sub-components with different continuity through learnable mapping and applies differentiated prediction, consistently maintaining the highest generation quality.

Furthermore, since the feature dynamics differ across timestep segments, LinCa trains isomorphic but separately parameterized predictors for each segment. Experiments show that partitioning into $S = 3$ segments already yields good results, and further increasing $S$ brings diminishing improvements. Detailed results, additional ablation studies, and more hyperparameter analyses are provided in the supplementary materials.

\section{Conclusion}

This paper proposes \textbf{\textit{LinCa}}, a feature caching acceleration framework based on learnable invertible networks. LinCa reveals that the hidden features of diffusion models exhibit significant dynamics differences across denoising stages, models, and feature dimensions. Based on it, LinCa decomposes cached features into sub-components with distinct continuity via a lightweight invertible network and applies differentiated polynomial prediction. Through data-driven per-segment training, LinCa adaptively accommodates the feature dynamics of different models and denoising stages, achieving a superior trade-off between acceleration and generation quality. Extensive experiments on FLUX, Qwen-Image, and HunyuanVideo demonstrate that LinCa significantly outperforms existing methods at the same acceleration ratio, achieving $5-7\times$ speedup with minimal quality degradation, while remaining compatible with distilled or quantized models. We believe LinCa opens up new possibilities for scalable, high-performance generative models and provides a new direction for efficient diffusion inference.

\section*{Acknowledgements}

This paper was partially sponsored by Terminal Intelligent Computing Division - Alibaba Cloud.

%
%

\bibliographystyle{splncs04}
\bibliography{main.bib}

\clearpage
\appendix
\section{Detailed Experiment Settings}

\textbf{Model Configurations and Evaluation Protocols.} We conduct comprehensive experiments covering seven generative models across three main tasks: text-to-image generation, text-to-video generation, and image editing. For text-to-image generation, FLUX.1-dev and Qwen-Image serve as our two primary evaluation targets. In addition, to verify that LinCa is compatible with mainstream model compression pipelines, we include three compressed derivatives of FLUX.1-dev: FLUX.1-lite-8B (model distillation), FLUX.1-schnell (step distillation), and FLUX.1-dev-int8 (quantization). For text-to-video generation, we adopt HunyuanVideo and evaluate it with VBench, which offers multi-dimensional quality assessments aligned with human perception. For image editing, Qwen-Image-Edit is benchmarked on GEdit-Bench, covering a wide range of editing types and prompt styles. Per-model details on output resolution, prompt selection, and evaluation metrics are described in the following subsections.

\subsection{Text-to-Image Generation}

\textbf{FLUX.1-dev}. Following the DrawBench evaluation protocol, we generate images conditioned on a fixed set of 200 prompts that cover a broad range of categories, including animals, colors, spatially conflicting descriptions, and fine-grained visual attributes. All images are produced at a resolution of $1024\times1024$ pixels, consistent with the model's recommended inference configuration. Generation quality is assessed using ImageReward, which captures photorealism and overall perceptual quality, alongside CLIP Score, which measures the degree of semantic alignment between the generated image and the input text prompt.

\textbf{Qwen-Image}. For Qwen-Image, we likewise adopt the DrawBench benchmark, using the same 200 prompts to ensure a fair comparison across models. Images are generated at a native resolution of $1328\times1328$ pixels. In addition to ImageReward and CLIP Score, which evaluate perceptual quality and text-image alignment respectively, we further report three pixel-level and perceptual fidelity metrics: Peak Signal-to-Noise Ratio (PSNR), Structural Similarity Index (SSIM), and Learned Perceptual Image Patch Similarity (LPIPS). These additional metrics provide a more complete assessment of how faithfully the accelerated model reproduces the output of the unaccelerated baseline.

\textbf{FLUX.1-lite-8B, FLUX.1-schnell, and FLUX.1-dev-int8.} To investigate whether LinCa can adapt to compressed models and deliver further acceleration while maintaining generation quality, we apply it to three compressed derivatives of FLUX.1-dev: FLUX.1-lite-8B (model distillation), FLUX.1-schnell (step distillation), and FLUX.1-dev-int8 (INT8 quantization). All three models are benchmarked on DrawBench with 200 prompts at $1024\times1024$ resolution. Despite their varied compression strategies, we apply a uniform evaluation protocol, measuring generation quality via ImageReward and CLIP Score, allowing us to quantify the additional benefit that LinCa provides when stacked on top of already-compressed models.

\subsection{Text-to-Video Generation}

\textbf{HunyuanVideo}. We benchmark HunyuanVideo on VBench, a human-aligned multi-dimensional evaluation suite designed specifically for video generation. The benchmark encompasses 946 prompts sourced from VBench-full-info.json, spanning a wide variety of semantic categories. Videos are generated at $480\times640$ resolution with 65 frames per clip, yielding rich temporal content for assessment. VBench evaluates generation quality along 18 fine-grained dimensions, covering aspects such as motion smoothness, visual coherence, temporal consistency, semantic alignment between the generated video and the text prompt, and overall perceptual quality.

\subsection{Image Editing}

\textbf{Qwen-Image-Edit}. We assess image editing performance on GEdit-Bench, a dedicated benchmark that contains 1212 editing prompts distributed across 11 editing categories, encompassing background replacement, color modification, material alteration, motion change, and pose adjustment, among others. Unlike image generation benchmarks, GEdit-Bench imposes no fixed output resolution, so images are generated without enforcing a specific spatial constraint. We report results on both the GEdit-CN and GEdit-EN subsets, evaluating each sample with three metrics: Overall Score (OS) for holistic editing performance, Perceptual Quality (PQ) for visual fidelity of the output, and Semantic Consistency (SC) for semantic alignment between the edited image and the editing instruction.

\subsection{Hardware}
All experiments are conducted on enterprise-grade GPU infrastructure:
\begin{itemize}[leftmargin=10pt,label=\textbf{•}]
    \item FLUX.1-dev experiments: NVIDIA A100 GPU
    \item FLUX.1-lite-8B experiments: NVIDIA A100 GPU
    \item FLUX.1-schnell experiments: NVIDIA A100 GPU
    \item FLUX.1-dev-int8 experiments: NVIDIA A100 GPU
    \item Qwen-Image experiments: NVIDIA H20 GPU  
    \item Qwen-Image-Edit experiments: NVIDIA H20 GPU  
    \item HunyuanVideo experiments: NVIDIA H100 GPU  
\end{itemize}

\subsection{Training Cost}

LinCa is trained only on pre-generated feature trajectories and does not require loading the diffusion model weights during training. In practice, 100--200 pre-generated feature samples are sufficient for each target model, and the training process can be completed within about one hour on a single GPU with 12GB VRAM. After training, the learned predictor is applied to unseen prompts and different caching intervals, which allows LinCa to provide a practical training-based acceleration module with modest one-time cost.

\section{Algorithmic Details}

\subsection{Caching Strategy and Interaction with Hyperparameter $N$}

The hyperparameter $N$ serves as the core control for the caching interval within LinCa. To address feature instability in the early denoising stages, we adopt a \texttt{first\_enhance} warmup stage, where the initial few timesteps are computed entirely by the diffusion backbone without any skipping. In the remaining stages, the denoising process alternates between computation steps (C) and prediction steps (P) at an interval of $N$: for every $N$ steps, one full forward pass is performed (C), and the subsequent $N{-}1$ steps are handled exclusively by the LinCa predictor (P) without invoking the diffusion backbone.

For example, in a 10-step diffusion process with \texttt{first\_enhance} = 3 and $N=3$, the execution sequence is \texttt{CCCPPCPPCP}. During prediction steps, the diffusion model is bypassed entirely. Instead, LinCa decomposes the most recently cached feature via the learned invertible mapping $\mathcal{E}_\theta^{(s)}$, applies differentiated-order prediction to each sub-component, and reconstructs the predicted feature through $(\mathcal{E}_\theta^{(s)})^{-1}$.

\textbf{Computation Steps.} At each full computation step at timestep $t$, we cache the cumulative residual feature $\mathbf{x}_t$ at the final layer of the diffusion model, and pass it through $\mathcal{E}_\theta^{(s)}$ to obtain the three sub-components:
\begin{equation}
  \mathbf{z}_t = \mathcal{E}_\theta^{(s)}(\mathbf{x}_t) = [\mathbf{z}_t^{(0)} \mid \mathbf{z}_t^{(1)} \mid \mathbf{z}_t^{(2)}].
\end{equation}
We additionally maintain the discrete differences of each sub-component to support higher-order prediction. For predicting step $t{-}k$, we cache the features from the three most recent computation steps $t$, $t+N$, and $t+2N$. Let $\mathbf{z}_{t}^{(m)}, \mathbf{z}_{t+N}^{(m)}, \mathbf{z}_{t+2N}^{(m)}$ denote the cached $m$-th sub-component at these three timesteps respectively. The first- and second-order discrete differences are computed as:
\begin{equation}
  \Delta^{(1)} \mathbf{z}^{(m)} = \mathbf{z}_{t}^{(m)} - \mathbf{z}_{t+N}^{(m)}, \qquad
  \Delta^{(2)} \mathbf{z}^{(m)} = \Delta^{(1)}\mathbf{z}^{(m)} - \left(\mathbf{z}_{t+N}^{(m)} - \mathbf{z}_{t+2N}^{(m)}\right).
\end{equation}

\textbf{Prediction Steps.} At each prediction step $t{-}k$ ($k \in \{1,\ldots,N{-}1\}$), LinCa applies differentiated-order Hermite extrapolation to each sub-component. Specifically, $\mathbf{z}^{(0)}$ is directly reused from the nearest computation step (0th-order), $\mathbf{z}^{(1)}$ is predicted via 1st-order extrapolation, and $\mathbf{z}^{(2)}$ via 2nd-order extrapolation:
\begin{equation}
  \hat{\mathbf{z}}_{t-k}^{(0)} = \mathbf{z}_{t}^{(0)}, \qquad
  \hat{\mathbf{z}}_{t-k}^{(m)} = \mathbf{z}_{t}^{(m)} + \sum_{i=1}^{m} \alpha_i(k)\, \Delta^{(i)} \mathbf{z}^{(m)}, \quad m \geq 1,
\end{equation}
where the Hermite interpolation coefficients $\alpha_i(k)$ are given by:
\begin{equation}
  \alpha_i(k) = \frac{H_i(x)}{i!} s^i, \qquad H_1(x) = 2x,\quad H_2(x) = 4x^2 - 2,\quad x = sk,
\end{equation}
with $s$ being a tuning factor and $k$ being the number of steps elapsed since the last computation step. The predicted sub-components are concatenated and reconstructed back to the original feature space via the inverse mapping:
\begin{equation}
  \hat{\mathbf{x}}_{t-k} = (\mathcal{E}_\theta^{(s)})^{-1}([\hat{\mathbf{z}}_{t-k}^{(0)} \mid \hat{\mathbf{z}}_{t-k}^{(1)} \mid \hat{\mathbf{z}}_{t-k}^{(2)}]).
\end{equation}

\subsection{Multi-Order Hermite Predictor and Efficient Computation}

To efficiently evaluate the Hermite prediction across all $N{-}1$ prediction steps within a caching interval, we employ a forward-difference operator $\Delta g(k)=g(k+1)-g(k)$. Substituting the Hermite coefficients into the prediction formulas, the explicit expressions for $\hat{\mathbf{z}}_{t-k}^{(1)}$ and $\hat{\mathbf{z}}_{t-k}^{(2)}$ are:
\begin{equation}
  \hat{\mathbf{z}}_{t-k}^{(1)} = \mathbf{z}_t^{(1)} + 2s^2k\,\Delta^{(1)}\mathbf{z}^{(1)}, \qquad
  \hat{\mathbf{z}}_{t-k}^{(2)} = \mathbf{z}_t^{(2)} + 2s^2k\,\Delta^{(1)}\mathbf{z}^{(2)} + \left(2s^4k^2 - s^2\right)\Delta^{(2)}\mathbf{z}^{(2)}.
\end{equation}
Since $\hat{\mathbf{z}}_{t-k}^{(1)}$ is linear in $k$ and $\hat{\mathbf{z}}_{t-k}^{(2)}$ is quadratic in $k$, their forward differences reduce to:
\begin{equation}
  \begin{aligned}
    \Delta\hat{\mathbf{z}}^{(1)}(k) &= 2s^{2}\,\Delta^{(1)}\mathbf{z}^{(1)}, \qquad
    \Delta^{m}\hat{\mathbf{z}}^{(1)}(k) = 0, \quad \text{for } m \geq 2, \\
    \Delta\hat{\mathbf{z}}^{(2)}(k) &= 2s^{2}\,\Delta^{(1)}\mathbf{z}^{(2)} + \left(4s^{4}k + 2s^{4}\right)\Delta^{(2)}\mathbf{z}^{(2)}, \\
    \Delta^{2}\hat{\mathbf{z}}^{(2)}(k) &= 4s^{4}\,\Delta^{(2)}\mathbf{z}^{(2)}, \qquad
    \Delta^{m}\hat{\mathbf{z}}^{(2)}(k) = 0, \quad \text{for } m \geq 3.
  \end{aligned}
\end{equation}
For $\mathbf{z}^{(1)}$, the forward difference $\Delta\hat{\mathbf{z}}^{(1)}$ is a constant, so each prediction step amounts to a single vector addition. For $\mathbf{z}^{(2)}$, $\Delta^{2}\hat{\mathbf{z}}^{(2)}$ is likewise computed once per caching interval, and $\Delta\hat{\mathbf{z}}^{(2)}(k)$ is updated recursively at each step. Together with the trivial reuse of $\mathbf{z}^{(0)}$, this closed-form recursive scheme ensures that the per-step overhead of the LinCa predictor remains negligible, regardless of the interval length $N$.

\subsection{Prediction Overhead}

Feature caching is effective because a cached prediction is substantially cheaper than a fresh diffusion forward pass. For FLUX.1-dev, one fresh denoising computation costs 74.39 TFLOPs, whereas one LinCa cached prediction costs only 0.02 TFLOPs. Therefore, replacing full computation steps with lightweight prediction steps brings significant acceleration while adding negligible predictor overhead.

\section{Mechanism}

\subsection{Continuity Disentanglement}

We further analyze why a learnable invertible mapping can separate features with different temporal continuity. Suppose the original feature can be written as $x=(x_{\rm dis},x_{\rm con})$, where the discontinuous component $x_{\rm dis}\in U_{\rm dis}$ and the continuous component $x_{\rm con}\in U_{\rm con}$ are interleaved in the original feature space. Let $\dim U_{\rm dis}=d_1$ and $\dim U_{\rm con}=d_2$. When $d_3\geq d_1$ and $d_4\geq d_2$, there exist injective mappings $A_{\rm dis}:U_{\rm dis}\hookrightarrow\mathbb{R}^{d_3}$ and $A_{\rm con}:U_{\rm con}\hookrightarrow\mathbb{R}^{d_4}$ such that the encoded feature can be represented as
\begin{equation}
  \mathcal{E}_{\theta}(x)=[z_{\rm dis},z_{\rm con}]
  =[A_{\rm dis}(x_{\rm dis}),A_{\rm con}(x_{\rm con})].
\end{equation}
Let $d=d_1+d_2$. Such a regrouping can be completed into an invertible transformation $W\in\mathbb{R}^{d\times d}$ satisfying $Wx=[z_{\rm dis},z_{\rm con}]$ and $\mathcal{E}_{\theta}^{-1}(z)=x$. This shows that the learnable invertible mapping used by LinCa contains feature-regrouping solutions that can disentangle continuous and discontinuous components into different subspaces. LinCa then applies order-matched prediction to these subspaces and reconstructs the final feature through the inverse mapping.

\subsection{Lossless Mapping}

The invertible projection in LinCa provides a strict reconstruction path from the decomposed feature space back to the original feature space. Each block consists of invertible operations, including the invertible $1\times1$ convolution and additive coupling layers, whose inverse transformations are explicitly defined. Therefore, the decomposition itself does not discard feature information. The prediction error mainly comes from estimating future sub-components, while the inverse mapping reconstructs the predicted feature back to the original space without introducing an additional projection bottleneck.

\subsection{Existing Methods}

\begin{figure}[H]
    \centering
    \includegraphics[width=0.5\linewidth]{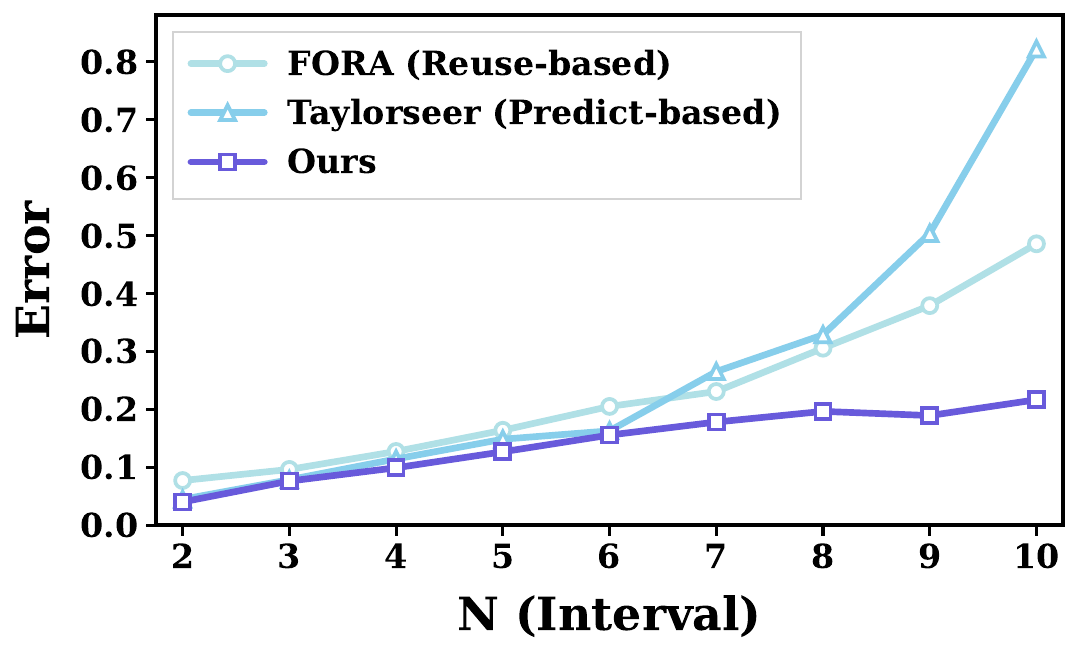}
    \caption{Prediction error between predicted and ground-truth features. LinCa obtains lower MSE by decomposing features into order-specific subspaces and applying matched prediction orders.}
    \label{fig:prediction_mse}
\end{figure}

Previous feature caching methods can be interpreted as special cases of LinCa. Reuse-based methods correspond to an identity mapping $\mathcal{E}_{0}:x\mapsto z=x$ followed by 0th-order prediction over all dimensions. Uniform prediction-based methods such as TaylorSeer also use the identity mapping, but apply the same high-order predictor to the full feature. In contrast, LinCa learns $\mathcal{E}_{\theta}:x\mapsto[z^{(0)},z^{(1)},z^{(2)}]$ and applies order-specific predictors to different subspaces, avoiding the limitation of imposing one prediction order on all dimensions.

Higher-order prediction is effective at relatively low acceleration ratios, but its accumulated error can become amplified under larger caching intervals. LinCa does not stack multiple predictors on the same full feature. Instead, it learns order-specific subspaces and applies the suitable predictor to each subspace. As shown in Figure~\ref{fig:prediction_mse}, this strategy yields lower prediction error than uniform prediction, especially when the acceleration ratio becomes larger.

\section{Hyperparameter Sensitivity Analyses}

\begin{figure}[H]
    \centering
    \includegraphics[width=1.0\linewidth]
    {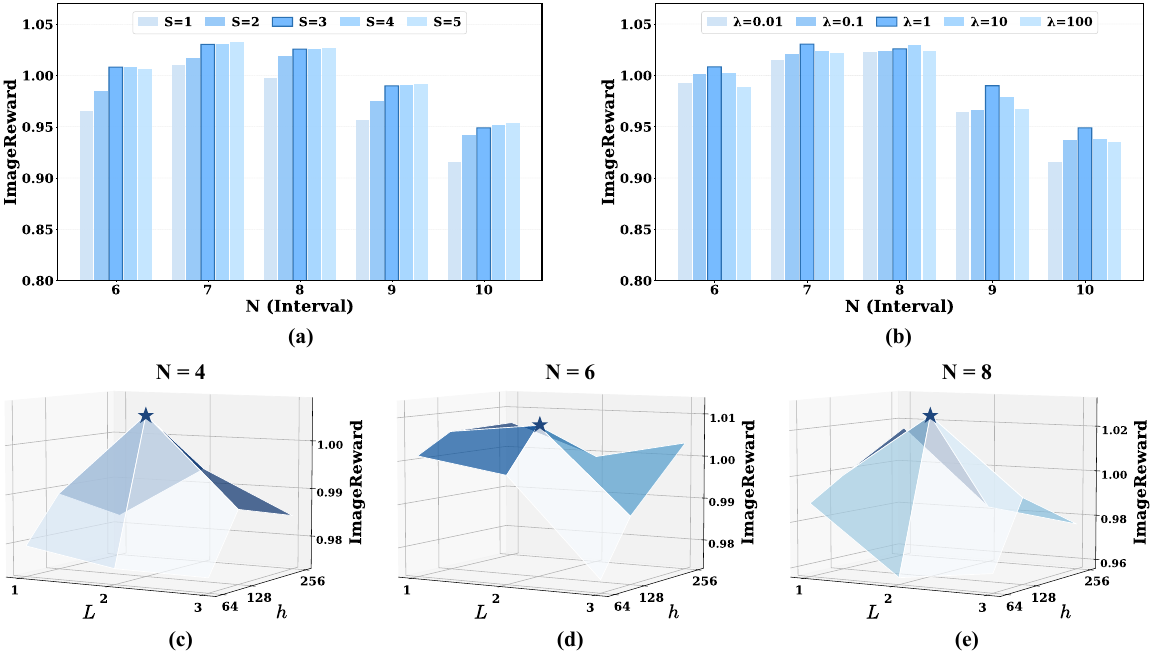}

    \captionof{figure}{\textbf{Hyperparameter sensitivity analyses of LinCa. (a)}: Impact of timestep segment number $S$. $S=3$ yields consistently strong performance, with further increases bringing only marginal gains. \textbf{(b)}: Impact of loss coefficient $\lambda$. $\lambda=1$ achieves the best performance across most intervals. \textbf{(c)-(e)}: Architecture of the decomposition network. $L=2$ invertible blocks with hidden dimension $h=128$ achieves the best performance across all intervals.}

    \label{hyperparameter}

\end{figure}

\subsection{Impact of Timestep Segment Number $S$}

We investigate how the number of timestep segments $S$ affects generation quality under varying caching intervals $N$. As shown in Figure~\ref{hyperparameter}(a), $S=1$ consistently yields the lowest performance across all intervals, while increasing $S$ leads to notable improvements, indicating that separate predictors for different denoising stages better accommodate their distinct feature dynamics. As $S$ increases from 1 to 3, performance improves consistently. However, as $S$ further increases to 4, 5, the additional gain becomes increasingly marginal. We therefore adopt $S=3$ as the default configuration in all experiments, balancing prediction fidelity with the modest overhead of maintaining separate predictor parameters per segment.

\subsection{Impact of Loss Coefficient $\lambda$}

We examine the sensitivity of LinCa to the loss balancing coefficient $\lambda$, which controls the relative weight of the sub-component prediction loss $\mathcal{L}_{\text{comp}}^{(s)}$ against the end-to-end feature prediction loss $\mathcal{L}_{\text{feat}}^{(s)}$ during training. As shown in Figure~\ref{hyperparameter}(b), $\lambda=1$ achieves the best performance across most caching intervals. When $\lambda$ is set too small (0.01 or 0.1), the sub-component supervision becomes insufficient, reducing the incentive for $\mathcal{E}_\theta^{(s)}$ to cluster dimensions according to their order-specific predictability. Conversely, when $\lambda$ is too large (10 or 100), the sub-component loss dominates the training objective and over-constrains the decomposition, interfering with the end-to-end reconstruction quality in the original feature space. A balanced $\lambda=1$ allows both losses to contribute constructively, and we adopt this value as the default configuration in all experiments.

\subsection{Architecture of the Decomposition Network}

We examine the effect of two key architectural hyperparameters of the learnable invertible network: the number of invertible blocks $L$ and the hidden dimension $h$. As shown in Figure~\ref{hyperparameter}(c)-(e), across all settings, $L=2$ with $h=128$ consistently achieves the best performance. Smaller configurations ($h=64$ or $L=1$) lack sufficient representational capacity to learn an effective invertible decomposition, resulting in consistently degraded prediction quality across all caching intervals. Larger configurations ($h=256$ or $L=3$), on the other hand, introduce excessive parameters without commensurate benefit, and the added complexity tends to destabilize training of the invertible network. We therefore adopt $L=2$ and $h=128$ as the default architecture.

\section{More Ablation Study}

\subsection{High-Order Predictor Choice}

\begin{table}[H]
\centering
\caption{Comparison of high-order predictors under similar acceleration ratios.}
\vspace{-3mm}
\begin{tabular}{c|c|c|c|c}
\toprule
\textbf{Method} & \textbf{Predictor} & \textbf{Speed$\uparrow$} & \textbf{ImageReward$\uparrow$} & \textbf{CLIP Score$\uparrow$} \\
\midrule
TaylorSeer      & Taylor    & 4.16$\times$ & 0.9793 & 32.63 \\
\midrule
\textbf{LinCa} & Hermite   & 5.51$\times$ & \textbf{1.0162} & \textbf{32.72} \\
\textbf{LinCa} & Taylor    & 5.51$\times$ & 1.0139 & 32.58 \\
\textbf{LinCa} & Chebyshev & 5.51$\times$ & 1.0023 & 32.60 \\
\textbf{LinCa} & Lagrange  & 5.51$\times$ & 1.0144 & 32.60 \\
\textbf{LinCa} & Laguerre  & 5.51$\times$ & 0.9264 & 32.50 \\
\bottomrule
\end{tabular}
\vspace{-3mm}
\label{tab:predictor_comparison}
\end{table}

We evaluate several candidate predictors for the higher-order sub-components of LinCa, including Hermite, Taylor, Chebyshev, Lagrange, and Laguerre interpolation. Multi-order Hermite interpolation consistently delivers the best generation quality and stability. Hermite benefits from incorporating both function values and derivative estimates through discrete differences, which enables it to capture the local curvature of the feature trajectory and maintain smooth transitions between prediction steps. This property is particularly important for the higher-order sub-components, which encode finer feature dynamics and are more sensitive to extrapolation errors at large caching intervals.

Taylor expansion tends to accumulate approximation error as the diffusion trajectory deviates from linearity, while Chebyshev, Lagrange and Laguerre polynomials introduce oscillatory behaviors that reduce prediction robustness. As shown in Table~\ref{tab:predictor_comparison}, at the same acceleration ratio ($5.51\times$), Hermite achieves an ImageReward of \textbf{1.0162} and CLIP Score of \textbf{32.72}, outperforming Lagrange (1.0144 / 32.60), Taylor (1.0139 / 32.58), Chebyshev (1.0023 / 32.60), and Laguerre (0.9264 / 32.50). Furthermore, applying Taylor directly to the full cached features as in TaylorSeer yields lower results even at a reduced acceleration ratio ($4.16\times$, 0.9793 / 32.63), further underscoring the value of learnable decomposition combined with a reliable multi-order Hermite predictor.

\section{More Experiments}

\subsection{Distillation and Training}

\begin{table}[H]
\centering
\caption{More results on FLUX.1-dev.}
\setlength{\tabcolsep}{3pt}
\resizebox{0.7\linewidth}{!}{%
\begin{tabular}{l|c|c|c}
\toprule
\textbf{Method} & \textbf{Steps} & \textbf{Speed $\uparrow$} & \textbf{ImageReward $\uparrow$} \\
\midrule
\textbf{FLUX.1-dev (Original)} & 50 & $1.00\times$ & 0.99 \\
\midrule
\multicolumn{4}{l}{\textit{Comparison with distilled models}} \\
\midrule
\textbf{FLUX.1-lite-8B} & 28 & $1.79\times$ & 0.89 \\
\rowcolor{gray}
\textbf{LinCa} $(N=4)$ & 14 & $\textbf{3.57}\times$ & \textbf{1.02} \\
\midrule
\rowcolor{white}
\multicolumn{4}{l}{\textit{Comparison with training-based methods}} \\
\midrule
\textbf{LESA} $(N=5)$ & 12 & $4.17\times$ & 1.00 \\
\rowcolor{gray}
\textbf{LinCa} $(N=6)$ & 10 & $\textbf{5.00}\times$ & \textbf{1.02} \\
\midrule
\rowcolor{white}
\textbf{LESA} $(N=7)$ & 9 & $5.56\times$ & 0.98 \\
\rowcolor{gray}
\textbf{LinCa} $(N=8)$ & 8 & $\textbf{6.25}\times$ & \textbf{1.02} \\
\bottomrule
\end{tabular}}
\label{tab:more_flux_results}
\end{table}

\begin{table}[H]
\centering
\caption{More results on Qwen-Image.}
\setlength{\tabcolsep}{3pt}
\resizebox{0.7\linewidth}{!}{%
\begin{tabular}{l|c|c|c}
\toprule
\textbf{Method} & \textbf{Steps} & \textbf{Speed $\uparrow$} & \textbf{ImageReward $\uparrow$} \\
\midrule
\textbf{Qwen-Image (Original)} & 50 & $1.00\times$ & 1.25 \\
\midrule
\multicolumn{4}{l}{\textit{Comparison with distilled models}} \\
\midrule
\textbf{Qwen-Image-Distill-Full} & 15 & $3.33\times$ & 1.02 \\
\textbf{Qwen-Image-Distill-LoRA} & 15 & $3.33\times$ & 0.95 \\
\rowcolor{gray}
\textbf{LinCa} $(N=4)$ & 14 & $\textbf{3.57}\times$ & \textbf{1.19} \\
\midrule
\rowcolor{white}
\multicolumn{4}{l}{\textit{Comparison with training-based methods}} \\
\midrule
\textbf{LESA} $(N=7)$ & 9 & $5.56\times$ & 1.15 \\
\rowcolor{gray}
\textbf{LinCa} $(N=7)$ & 9 & $5.56\times$ & \textbf{1.17} \\
\midrule
\rowcolor{white}
\textbf{LESA} $(N=10)$ & 7 & $7.14\times$ & 1.01 \\
\rowcolor{gray}
\textbf{LinCa} $(N=10)$ & 7 & $7.14\times$ & \textbf{1.05} \\
\bottomrule
\end{tabular}}
\label{tab:more_qwen_results}
\end{table}

We provide more comparisons with distilled models and training-based acceleration methods in Table~\ref{tab:more_flux_results} and Table~\ref{tab:more_qwen_results}. On FLUX.1-dev, LinCa achieves a better speed-quality trade-off than FLUX.1-lite-8B and LESA. On Qwen-Image, LinCa also outperforms Qwen-Image-Distill-Full, Qwen-Image-Distill-LoRA, and LESA under comparable or higher acceleration ratios. These results show that LinCa is complementary to model compression and competitive with training-based acceleration methods.

In some settings, accelerated variants can slightly exceed the original model under preference-based metrics. This behavior suggests that feature prediction may impose mild temporal regularization on cached representations. Similar observations have also been reported in cache-based methods such as TaylorSeer and FoCa. Therefore, we report both preference-based metrics and fidelity metrics whenever applicable.

\subsection{Few-Step Distillation}

\begin{table}[H]
\centering
\caption{Compatibility with Qwen-Image-Lightning.}
\setlength{\tabcolsep}{3pt}
\resizebox{0.7\linewidth}{!}{%
\begin{tabular}{l|c|c|c|c}
\toprule
\textbf{Method} & \textbf{Steps} & \textbf{Speed $\uparrow$} & \textbf{ImageReward $\uparrow$} & \textbf{GenEval $\uparrow$} \\
\midrule
\textbf{Qwen-Image-Lightning} & 8 & $1.00\times$ & 1.28 & 0.84 \\
\rowcolor{gray}
\textbf{LinCa} $(N=3)$ & 4 & $2.00\times$ & 1.26 & 0.85 \\
\bottomrule
\end{tabular}}
\label{tab:qwen_lightning}
\end{table}

Table~\ref{tab:qwen_lightning} further evaluates LinCa on Qwen-Image-Lightning. Feature caching is mainly applicable to multi-step denoising models pursuing high quality, and the exploitable redundancy becomes limited in extremely few-step settings. Nevertheless, for the 8-step Qwen-Image-Lightning model, LinCa still provides an additional $2.00\times$ acceleration while maintaining comparable ImageReward and slightly improving GenEval.

\subsection{Wan2.1}

\begin{table}[H]
\centering
\caption{Text-to-video generation result on Wan2.1.}
\setlength{\tabcolsep}{3pt}
\resizebox{0.7\linewidth}{!}{%
\begin{tabular}{l|c|c|c}
\toprule
\textbf{Method} & \textbf{Steps} & \textbf{Speed $\uparrow$} & \textbf{VBench Score(\%) $\uparrow$} \\
\midrule
\textbf{Wan2.1-1.3B (Original)} & 50 & $1.00\times$ & 83.72 \\
\midrule
\textbf{TaylorSeer} $(N=5)$ & 12 & $4.17\times$ & 81.07 \\
\rowcolor{gray}
\textbf{LinCa} $(N=6)$ & 10 & $\textbf{5.00}\times$ & \textbf{82.56} \\
\bottomrule
\end{tabular}}
\label{tab:wan21}
\end{table}

We also evaluate LinCa on the stronger Wan2.1-1.3B video generation model. As shown in Table~\ref{tab:wan21}, LinCa outperforms TaylorSeer in both acceleration ratio and VBench score, demonstrating its generalization to another video diffusion model. For HunyuanVideo in our paper, we follow TaylorSeer and FoCa at $480\times640$ resolution for fair comparison, which differs from the $720\times1280$ leaderboard setting.

\end{document}